\documentclass[10pt,twocolumn,letterpaper]{article}

\usepackage[pagenumbers]{cvpr} % To force page numbers, e.g. for an arXiv version

\usepackage{longtable}
\usepackage{multirow}
\usepackage{algorithm}
\usepackage{algpseudocode}
\usepackage{amsmath}
\usepackage{amsthm}

\usepackage{changes}
\definechangesauthor[name=Runze, color=orange]{wrz}
\usepackage{todonotes}

\definecolor{cvprblue}{rgb}{0.21,0.49,0.74}
\usepackage[pagebackref,breaklinks,colorlinks,allcolors=cvprblue]{hyperref}

\def\paperID{xxxx} % *** Enter the Paper ID here
\def\confName{CVPR}
\def\confYear{2026}

\title{STAC: Plug-and-Play Spatio-Temporal Aware Cache Compression for Streaming 3D Reconstruction}

\author{
Runze Wang \quad
Yuxuan Song \quad
Youcheng Cai\thanks{Corresponding author.} \quad
Ligang Liu \\
University of Science and Technology of China
\\ {\tt\scriptsize runzewang@mail.ustc.edu.cn \quad syx121900@mail.ustc.edu.cn \quad caiyoucheng@ustc.edu.cn \quad lgliu@ustc.edu.cn}
}

\begin{document}
\maketitle
\begin{abstract}
Online 3D reconstruction from streaming inputs requires both long-term temporal consistency and efficient memory usage. Although causal variants of VGGT address this challenge through a key-value (KV) cache mechanism, the cache grows linearly with the stream length, creating a major memory bottleneck. Under limited memory budgets, early cache eviction significantly degrades reconstruction quality and temporal consistency. In this work, we observe that attention in causal transformers for 3D reconstruction exhibits intrinsic spatio-temporal sparsity. Based on this insight, we propose \textbf{STAC}, a \textbf{S}patio-\textbf{T}emporally \textbf{A}ware \textbf{C}ache Compression framework for streaming 3D reconstruction with large causal transformers. STAC consists of three key components: (1) a \textbf{Working Temporal Token Caching} mechanism that preserves long-term informative tokens using decayed cumulative attention scores; (2) a \textbf{Long-term Spatial Token Caching} scheme that compresses spatially redundant tokens into voxel-aligned representations for memory-efficient storage; and (3) a \textbf{Chunk-based Multi-frame Optimization} strategy that jointly processes consecutive frames to improve temporal coherence and GPU efficiency. Extensive experiments show that STAC achieves state-of-the-art reconstruction quality while reducing memory consumption by nearly 10$\times$ and accelerating inference by 4$\times$, substantially improving the scalability of real-time 3D reconstruction in streaming settings. 
\textit{Project page:} \href{https://stac-3r.github.io/}{https://stac-3r.github.io/}.
\end{abstract}    
\section{Introduction}
\begin{figure}[t]
    \centering
    \includegraphics[width=0.935\linewidth]{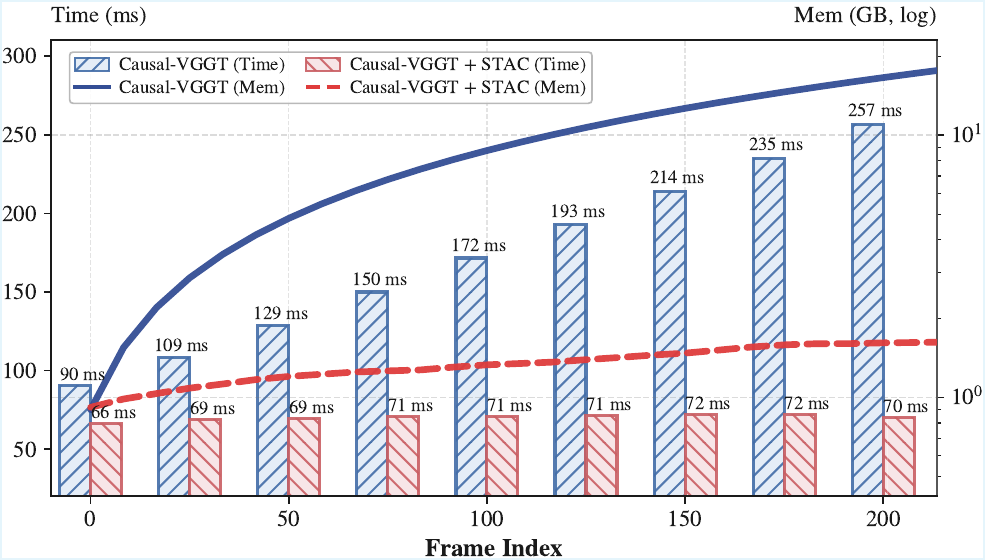}
    % \caption{STAC explicitly models both spatial and temporal sparsity for efficient and consistent 3D reconstruction in streaming scenarios. The bottom comparison illustrates the superior memory efficiency and runtime performance of our method.}
    \caption{\textbf{Runtime--memory scaling in streaming 3D reconstruction.}  Bars show per-frame runtime (ms) and lines show KV cache memory (GB, log scale) as the stream grows. Compared with Causal-VGGT, \textbf{STAC} reduces KV cache growth and stabilizes per-frame runtime as the stream length increases.}
    \label{fig:teaser}
    \vspace{-16pt}
\end{figure}
Reconstructing dense 3D geometry from images remains a fundamental problem in computer vision and graphics. It supports a wide range of applications in virtual and augmented reality~\cite{jiang2024vr, sakr2024virtual}, robotics~\cite{wang2024large}, and autonomous driving~\cite{reda2024path}. Traditional reconstruction pipelines such as Structure from Motion (SfM)~\cite{schonberger2016structure} and Multi-View Stereo (MVS)~\cite{furukawa2015multi} rely on explicit geometric constraints and optimization to estimate geometry and camera motion. Although these methods are well-established, their high computational cost and limited scalability restrict their applicability in real-time and large-scale scenarios.

Recently, transformer-based architectures are reshaping the field of 3D geometry estimation. VGGT~\cite{wang2025vggt} introduces a feed-forward transformer that jointly infers camera parameters, depth maps, point maps, and 3D point tracks from an arbitrary number of input views, achieving state-of-the-art performance without external geometric optimization. Despite this progress, existing methods typically rely on global self-attention or full-sequence processing. In streaming or incremental scenarios, such architectures require all frames to be available beforehand or necessitate recomputation over previous frames, making this batch-processing paradigm unsuitable for online 3D reconstruction and leading to memory and latency inefficiencies. Recent causal variants of VGGT (Causal-VGGT), such as StreamVGGT~\cite{zhuo2025streaming} and STream3R~\cite{lan2025stream3r}, alleviate these issues by replacing global self-attention with causal self-attention, thereby enabling streaming reconstruction, yet their key-value (KV) cache still grows linearly with sequence length, resulting in substantial memory consumption and increased latency.

Through analysis of the Causal-VGGT architecture, we observe that its KV cache exhibits structured sparsity along two complementary dimensions: \emph{spatial sparsity}, where tokens correlate with 3D positions and viewpoint changes, and \emph{temporal sparsity}, where certain tokens persistently contribute across frames. However, existing streaming-based approaches treat tokens uniformly, ignoring this structured sparsity, leading to premature eviction of informative tokens and unnecessary caching of transient ones.

% To address these limitations, we propose \textbf{STAC}, a plug-and-play framework for causal transformer-based 3D reconstruction that models structured spatio-temporal sparsity within the KV cache (see Fig. \ref{fig:teaser}). Our framework explores \textit{spatio-temporal KV cache compression} for Causal-VGGT without additional training, enabling long-term reuse of informative tokens in a training-free and memory-efficient manner. First, we introduce a \textbf{Working Temporal Token Caching} mechanism that maintains a dynamic temporal cache consisting of global reference, local window, and anchor tokens. This cache exploits temporal persistence by adaptively updating token importance through decayed cumulative attention, ensuring long-term coherence during streaming reconstruction. Second, we design a \textbf{Long-term Spatial Token Caching} scheme that leverages the spatial redundancy of 3D scenes. Evicted tokens are organized into voxel-based structures and compressed through similarity-weighted merging, yielding compact yet informative voxel-level representations that preserve geometric details for future reuse. Finally, we propose a \textbf{Chunk-based Multi-frame Optimization} framework that groups consecutive frames into temporal chunks for joint refinement. This strategy enhances temporal consistency and geometric accuracy while exploiting GPU parallelism for low-latency inference. 

To address these limitations, we propose \textbf{STAC}, a plug-and-play framework for causal transformer-based 3D reconstruction that exploits structured spatio-temporal sparsity in the KV cache. Without additional training, STAC performs spatio-temporal KV cache compression for Causal-VGGT, enabling long-term reuse of informative tokens in a memory-efficient manner.
Inspired by mechanisms of human memory, STAC maintains two complementary forms of memory. 
First, \textbf{Working Temporal Token Caching} maintains a short-term, high-fidelity working memory that retains recent observations together with a small set of persistent anchor tokens. It combines global reference tokens, a sliding local window, and dynamically selected anchors whose importance is updated via decayed cumulative attention, capturing short-term continuity while preserving globally important cues.
% Second, \textbf{Long-term Spatial Token Caching} maintains a geometry-aware long-term memory. Instead of storing all historical tokens explicitly, it organizes evicted tokens within a voxel grid and progressively merges them into compact voxel-level representations, preserving early-observed geometric information while bounding memory growth over long sequences.
Second, \textbf{Long-term Spatial Token Caching} maintains a geometry-aware long-term memory. Instead of storing all historical tokens explicitly, it organizes evicted tokens within a voxel grid and progressively merges them into compact voxel-level representations, enabling efficient spatial retrieval of relevant tokens while preserving early-observed geometric information and bounding memory growth over long sequences.
Finally, we introduce \textbf{Chunk-based Multi-frame Optimization}, which groups a small number of already arrived frames into temporal chunks for joint processing. This allows limited intra-chunk information sharing without accessing future frames, improving both reconstruction consistency and GPU utilization in the streaming setting.

Our main contributions are summarized as follows:
\begin{itemize}
\item {We identify and analyze structured spatio-temporal sparsity in causal-transformer 3D reconstruction, laying the foundation for a unified cache compression framework.}
    % \item {We propose the spatio-temporal aware cache compression module, consisting of working temporal caching, long-term spatial caching, and efficient token retrieval to enable compact and consistent memory management.}
    \item We propose a spatio-temporal aware cache compression module that integrates working temporal caching and long-term spatial caching for compact and consistent memory management in streaming 3D reconstruction.
    \item {We introduce a chunk-based multi-frame strategy that jointly refines consecutive frames, improving temporal coherence and leveraging hardware parallelism.}
\end{itemize}
% Our main contributions are summarized as follows:
% \begin{itemize}
%     \item {We identify and analyze the structured spatio-temporal sparsity in causal transformer-based 3D reconstruction, motivating a training-free spatio-temporal KV cache compression framework.}
%     \item {We propose STAC, a plug-and-play spatio-temporal KV cache compression framework with two complementary mechanisms: Working Temporal Token Caching for informative temporal memory, and Long-term Spatial Token Caching for compact geometry-aware long-term memory.}
%     \item {We further introduce Chunk-based Multi-frame Optimization, which jointly processes consecutive arrived frames to improve temporal coherence and GPU efficiency in streaming reconstruction.}
% \end{itemize}

To the best of our knowledge, STAC provides the first systematic study of \textit{training-free spatio-temporal KV cache compression} for causal transformer-based 3D reconstruction. Across standard benchmarks, STAC reduces memory usage by nearly 10$\times$ and achieves a 4$\times$ inference speed-up, while delivering reconstruction quality that remains virtually indistinguishable from full Causal-VGGT models.

\section{Related Work}
\subsection{Image-based 3D Reconstruction}
% \,\,\,\,  \textbf{Traditional approaches.} Traditional 3D reconstruction from image collections is typically optimization-based and tailored to specific scenes. Structure-from-Motion (SfM) \cite{schonberger2016structure, sweeney2015optimizing, wilson2014robust, wu2013towards} performs feature extraction and matching \cite{lowe2004distinctive, dusmanu2019d2, sarlin2020superglue, chen2021learning}, triangulation, and bundle adjustment \cite{triggs1999bundle, agarwal2010bundle}. Dense reconstruction can be obtained through Multi-View Stereo (MVS) \cite{furukawa2015multi, schonberger2016pixelwise, vats2024gc}. While these modular pipelines demonstrate strong accuracy and robustness, they require sequential optimization, are computationally expensive, and are highly sensitive to noise, occlusion, and viewpoint variations.
\,\,\,\, \textbf{Traditional methods.}
Classical 3D reconstruction pipelines rely on optimization-based, scene-specific processing. Structure-from-Motion (SfM) ~\cite{schonberger2016structure, sweeney2015optimizing, wilson2014robust, wu2013towards} estimates camera poses via feature extraction and matching ~\cite{lowe2004distinctive, dusmanu2019d2, sarlin2020superglue, chen2021learning}, triangulation, and bundle adjustment ~\cite{triggs1999bundle, agarwal2010bundle}, while Multi-View Stereo (MVS) ~\cite{furukawa2015multi, schonberger2016pixelwise, vats2024gc} recovers dense geometry.
Although accurate and robust, these modular pipelines require sequential optimization, are computationally expensive, and remain sensitive to noise, occlusion, and viewpoint changes.

% \textbf{Offline approaches.} Recent research has shifted toward feed-forward, end-to-end 3D reconstruction frameworks that bypass explicit geometric optimization. DUSt3R~\cite{wang2024dust3r} formulates dense reconstruction as direct pointmap regression, avoiding dependence on known camera intrinsics or poses. MASt3R~\cite{leroy2024grounding} extends this formulation by grounding correspondences in 3D space to improve generalization and robustness. VGGT \cite{wang2025vggt} further unifies multi-view geometry by jointly predicting camera parameters, depth, pointmaps, and feature tracks within a single transformer-based architecture. Subsequent variants—such as Fast3R \cite{yang2025fast3r}, VGGT-Long \cite{deng2025vggt}, and FastVGGT \cite{shen2025fastvggt}—scale these architectures to thousands of images and kilometer-scale sequences through parallelization or training-free token merging. However, these models rely on full self-attention across all input tokens, resulting in quadratic computational and memory complexity. Consequently, they remain inherently \emph{offline}, requiring complete re-inference when new frames are introduced.
\textbf{Offline approaches.}
Recent work has shifted toward feed-forward, end-to-end 3D reconstruction frameworks that bypass explicit geometric optimization. DUSt3R~\cite{wang2024dust3r} models dense reconstruction as direct pointmap regression without requiring camera intrinsics or poses, while MASt3R~\cite{leroy2024grounding} grounds correspondences in 3D space to enhance generalization. Visual Geometry Grounded Transformer (VGGT)~\cite{wang2025vggt} further unifies multi-view geometry by jointly predicting camera parameters, depth, pointmaps, and feature tracks within a single transformer architecture.
Recent variants~\cite{yang2025fast3r,deng2025vggt,shen2025fastvggt} scale these models to thousands of images and kilometer-scale sequences via parallelization or training-free token merging. Still, they rely on full self-attention over all input tokens, incurring quadratic compute and memory costs and requiring re-inference when new frames arrive—limiting their use in streaming scenarios.

\textbf{Online/Streaming approaches.}
To enable real-time and scalable 3D reconstruction, recent works have explored causal and online transformer formulations. Building on the DUSt3R paradigm, prior studies~\cite{wang2025continuous,wang2024spann3r,wu2025point3r} introduce memory mechanisms—such as persistent latent tokens, spatial feature banks, and 3D-aware pointer memories—to support online reconstruction from image streams, but often require architectural specialization or task-specific fine-tuning.
More recently, StreamVGGT~\cite{zhuo2025streaming} and STream3R~\cite{lan2025stream3r} adapt the bidirectional VGGT~\cite{wang2025vggt} to a causal setting (Causal-VGGT). They convert the backbone into a causal transformer and maintain a growing KV cache over past frames to provide temporal context during streaming inference. Benefiting from VGGT’s strong representational capacity, STream3R~\cite{lan2025stream3r} in particular achieves state-of-the-art performance on online 3D reconstruction benchmarks.
Despite these advances, existing Causal-VGGT methods still suffer from memory and latency inefficiencies: the KV cache grows linearly with sequence length, leading to high memory consumption and slower inference over long or continuous streams. In contrast, our method introduces a spatio-temporal aware cache compression framework that compresses redundant historical states, maintaining bounded memory usage with minimal impact on reconstruction accuracy.

\subsection{KV Cache Compression}
KV cache compression has been extensively studied in LLMs to alleviate memory and latency bottlenecks under long-context inference. Eviction-based methods such as H2O~\cite{zhang2023h2o} and StreamLLM~\cite{xiao2023efficient} retain only a small set of high-importance tokens while discarding the rest, achieving substantial compression with limited quality loss. Merge-based approaches~\cite{wan2024d2o,wang2024model} further exploit redundancy in sequential inputs by merging tokens that are about to be evicted into nearby important tokens, thereby reducing information loss under tight memory budgets. In multimodal and streaming settings, frameworks~\cite{di2025streaming,yang2025streammem,ning2025livevlm} adopt query- or saliency-driven compression and retrieval, but mainly leverage similarity across neighboring frames or spatial regions. In contrast, our work targets streaming 3D reconstruction and, to our knowledge, is the first to reveal strong spatio-temporal sparsity and local spatial redundancy in the KV cache of large transformer-based 3D models. We explicitly exploit this structure to design a causal, spatio-temporal aware KV compression scheme whose representations remain compact yet temporally coherent, thereby supporting continuous and accurate {3D perception.}

\section{Preliminaries}
% \,\,\,\, \textbf{VGGT.} The Visual Geometry Grounded Transformer (VGGT) \cite{wang2025vggt} is a unified, transformer-based framework designed to reconstruct 3D scene geometry from multi-view images. Given a sequence of frames $\{I_t\}_{\tau=1}^{T}$ captured from a 3D scene, VGGT first employs an image encoder to transform each frame $I_t$ into a set of visual tokens $\{F_t\}_{t=1}^{T}$. These tokens serve as compact feature representations for each individual view. Subsequently, a multi-view decoder aggregates cross-frame information through global self-attention, effectively reasoning across spatial and viewpoint variations to produce geometry-aware tokens $\{G_t\}_{t=1}^{T}$. Finally, per-view prediction heads decode these geometry-aware tokens to estimate 3D outputs, including point maps $P_t$, depth maps $D_t$, and camera parameters $c_t$. The overall pipeline can be formulated as follows:

% \begin{equation}
% \begin{gathered}
% F_t = \operatorname{Encoder}(I_t), \\
% \{G_t\}_{t=1}^{T} = \operatorname{Decoder}\left(\operatorname{Global\, SelfAttn}(\{F_t\}_{t=1}^{T})\right) \\
% c_t, D_t, P_t = \operatorname{DecodingHeads}(G_t)    
% \end{gathered}
% \end{equation}

\,\,\,\, \textbf{VGGT.} The Visual Geometry Grounded Transformer (VGGT)~\cite{wang2025vggt} is a unified transformer-based framework for 3D reconstruction from multi-view images. Given a batch of input frames, an image encoder~\cite{oquab2023dinov2} extracts corresponding visual tokens $F_t$ for each frame. These tokens are jointly aggregated using global self-attention within a multi-view decoder to produce geometry-aware representations:
\begin{equation}
\{G_t\}_{t=1}^{T} = \operatorname{Decoder}\left(\operatorname{Global\, SelfAttn}(\{F_t\}_{t=1}^{T})\right).
\end{equation}

The geometry-aware tokens $G_t$ are passed to prediction heads~\cite{ranftl2021vision} to estimate per-frame camera parameters $c_t$, depth maps $D_t$, and point maps $P_t$:
\begin{equation}
c_t, D_t, P_t = \operatorname{HeadDecoder}(G_t).
\end{equation}

\textbf{Causal-VGGT}. While VGGT operates offline by jointly processing all frames, Causal-VGGT~\cite{zhuo2025streaming, lan2025stream3r} extends the architecture to support streaming inference for 3D reconstruction. It replaces the global self-attention module in the decoder with a causal self-attention mechanism that attends only to past frames, enforcing temporal consistency.

To support this causal computation efficiently, Causal-VGGT maintains a key-value (KV) cache that accumulates attention keys and values from previous frames across all decoder layers and attention heads~\cite{vaswani2017attention,brown2020language}. Specifically, at time step $t$, each layer $\ell$ and head $h$ stores:
\begin{equation}
\mathcal{M}_t^{\ell,h} = \{\mathbf m_\tau^{\ell,h} \}_{\tau=1}^{t-1} = \{ \mathbf k_\tau^{\ell,h}, \mathbf v_\tau^{\ell,h} \}_{\tau=1}^{t-1},
\end{equation}
where $\mathbf{k}_\tau^{\ell,h}$ and $\mathbf{v}_\tau^{\ell,h}$ are computed via learned projections at each previous step $\tau$. These cached representations allow the decoder to incorporate long-term temporal context without reprocessing prior frames.

At each time step, the decoder applies causal self-attention to the current frame's features $F_t$, using the cached keys and values to incorporate temporal context:
\begin{equation}
G_t = \operatorname{Decoder}( \operatorname{Causal\,SelfAttn}(F_t, \{ \mathcal{M}_t^{\ell,h} \}_{\ell,h}) ).
\label{eq:crossattn}
\end{equation}

This causal formulation enables efficient streaming inference while maintaining temporal coherence and long-range dependencies learned during training.

\section{Observation}
\label{sec:observation}

\begin{figure}[t]
    \centering
    \includegraphics[width=0.9\linewidth]{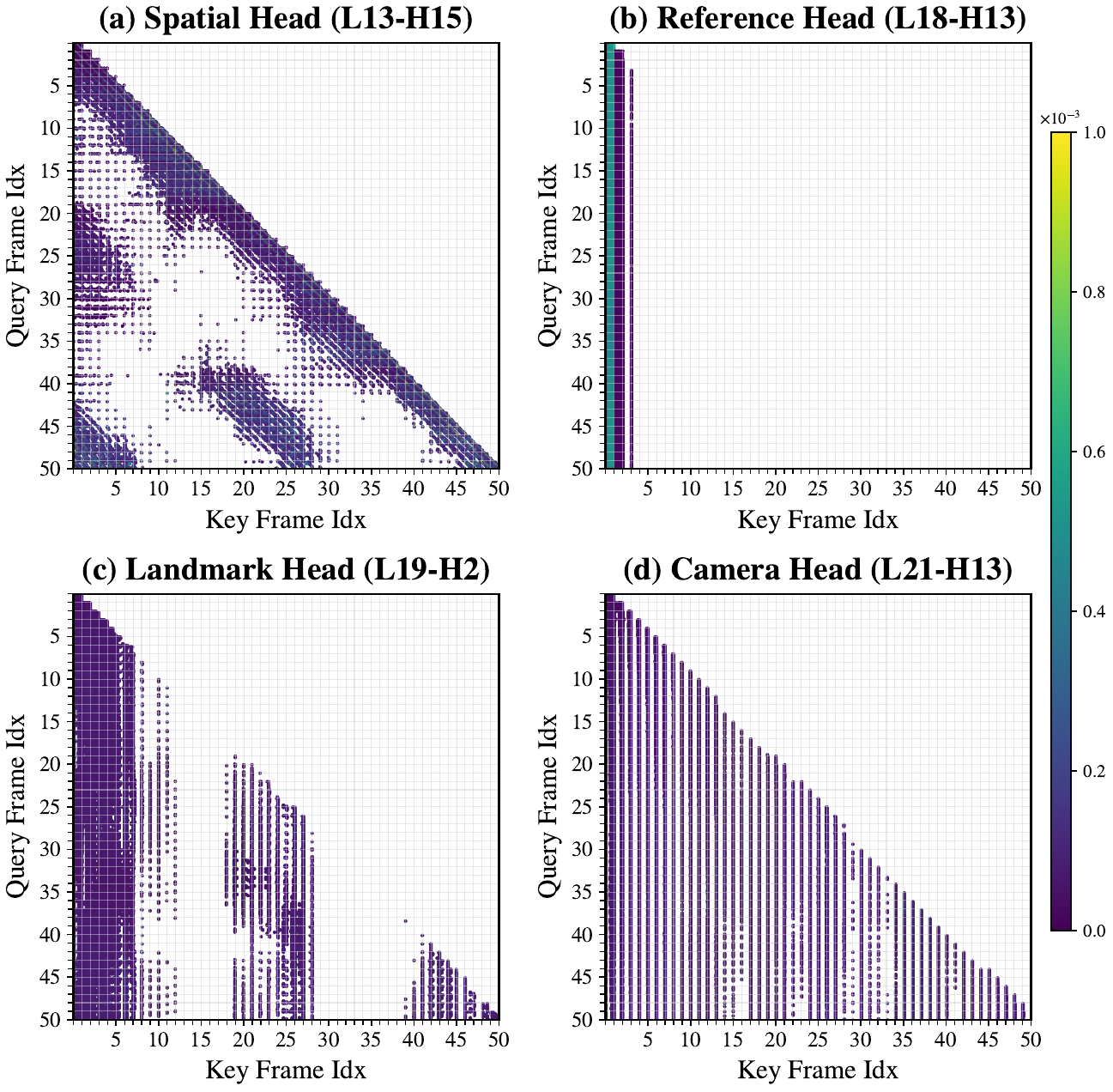}
    \caption{\textbf{Spatio-temporal attention sparsity.} Representative attention patterns in the global causal attention map of Causal-VGGT, {retaining the top 1024 keys per query for visualization}. (a) Spatial attention aligned with camera motion; (b) Persistent focus on first-frame tokens as global references; (c) Temporal anchoring via tokens from semantically stable {landmark frames};  (d) Long-range attention to camera tokens encoding global context.}
    \label{fig:patterns}
    \vspace{-12pt}
\end{figure}

We begin by analyzing the Causal-VGGT architecture and identify two distinct forms of sparsity within its KV cache: \emph{spatial} and \emph{temporal} sparsity. Visualizations of attention maps (Fig.~\ref{fig:patterns}) reveal that different heads specialize in distinct roles—some focus on spatial reasoning, others on temporal consistency. This indicates that Causal-VGGT implicitly organizes spatio-temporal structure, leading to differentiated sparsity patterns in its internal representation.

% By systematically visualizing attention maps across layers and heads, we observe that different attention heads specialize in complementary roles, some emphasizing spatial reasoning, while others focus on temporal association. Collectively, these patterns reveal that VGGT implicitly organizes spatio-temporal information, thereby exposing the inherent structure of spatial and temporal sparsity within its internal representations.

% \textbf{Spatial Sparsity.} A subset of attention heads exhibits pronounced spatial sensitivity that is strongly correlated with camera motion. In 3D reconstruction tasks, where cameras frequently revisit under-reconstructed regions or follow loop-closure trajectories, these heads exhibit spatially coherent attention: query states tend to focus on visually similar and spatially adjacent regions across neighboring frames (see Fig.~\ref{fig:spati-temp}). This behavior suggests that these heads implicitly encode spatial priors, thereby facilitating local feature matching and temporal alignment. Consequently, they enhance 3D feature aggregation, reinforce cross-view correspondence, and improve structural consistency under complex camera motions. Because tokens associated with these spatial heads are highly sensitive to viewpoint changes, their representations are likely correlated with the underlying 3D spatial positions. We validate this hypothesis by analyzing the relationship between token feature similarity and the corresponding 3D coordinates (see Appendix A.1).

\textbf{Spatial Sparsity.} Certain attention heads exhibit strong spatial sensitivity correlated with camera motion. During loop closures or revisits, these heads consistently focus on visually and spatially adjacent regions across frames (Fig.~\ref{fig:patterns}a), indicating that they capture spatial priors useful for cross-view matching and structural consistency. We validate this behavior by analyzing feature similarity and corresponding 3D coordinates, showing that view-dependent tokens often exhibit redundancy in spatially adjacent regions (see Supplementary Material, Sec.~A).

% \textbf{Temporal Sparsity.} 
% Another group of heads exhibits strong temporal sensitivity, persistently attending to a specific subset of tokens over time. We identify three major temporal correlation patterns: (1) \textit{First-frame correlation}: certain heads attend to tokens from the first frame (Fig.~\ref{fig:first}), indicating the model’s reliance on early observations as a global reference frame. In streaming reconstruction, this frame serves as the spatial origin that anchors subsequent frames and ensures global consistency. (2) \textit{Key-frame correlation}: certain heads maintain high attention toward key frames containing semantically dominant regions. These tokens act as stable visual anchors that preserve consistency during long-term streaming reconstruction. (3) \textit{Camera-token correlation}: several heads exhibit persistent attention to camera tokens (Fig.~\ref{fig:cam-massive}), which propagate global contextual cues and regulate downstream token embeddings across the sequence.
\textbf{Temporal Sparsity.} Other heads display persistent attention to a small subset of tokens over time, revealing three patterns:  
(1) \textit{First-frame correlation} – some heads repeatedly attend to the first frame as a global reference (Fig.~\ref{fig:patterns}b);
(2) \textit{Landmark-frame correlation} – others consistently attend to tokens from semantically stable landmark frames, serving as long-term temporal cues (Fig.~\ref{fig:patterns}c);
(3) \textit{Camera-token correlation} – several heads attend to camera tokens, propagating global context (Fig.~\ref{fig:patterns}d).

Despite these structured sparsity patterns, prior streaming 3D methods~\cite{zhuo2025streaming, lan2025stream3r} ignore such distinctions and treat all tokens equally. Uniform caching/eviction policies often discard globally important tokens prematurely. This motivates the design of our \textbf{Spatio-Temporal Aware Cache Compression (STAC)} module, which explicitly models these sparsity patterns by preserving informative tokens while compactly storing view-dependent ones for future retrieval. This design improves both efficiency and temporal consistency in streaming 3D reconstruction.

\begin{figure*}[t]
    \centering
    \includegraphics[width=\textwidth]{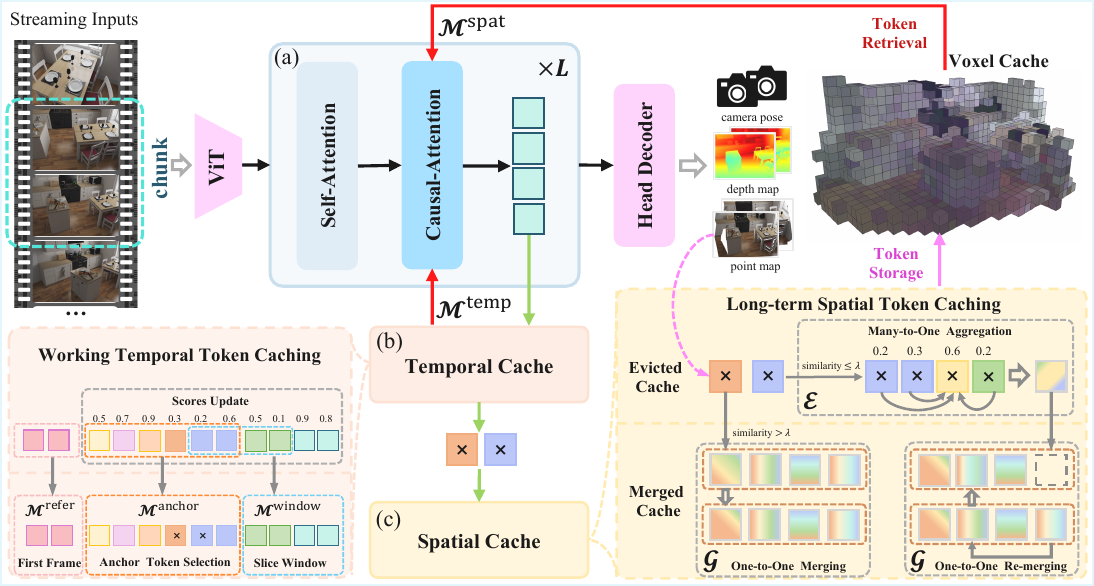}
% \caption{\textbf{Overview of STAC.} Our framework reconstructs 3D scenes online from a video stream via spatio-temporal token caching with \emph{chunk-based} causal inference.
% (a) The \textbf{Causal-VGGT} module processes ViT-tokenized frames in a chunk with global causal attention over the working temporal cache $\mathcal M^{\text{temp}}$ and voxel-retrieved spatial tokens $\mathcal M^{\text{spat}}$. 
% (b) During inference, \textbf{Working Temporal Token Caching} updates token scores after each KV cache access, retaining high-scoring anchors $\mathcal M^{\text{anchor}}$ while always preserving first-frame tokens $\mathcal M^{\text{refer}}$ and sliding-window tokens $\mathcal M^{\text{window}}$, and evicting the rest. 
% (c) Finally, the \textbf{Long-term Spatial Token Caching} scheme routes evicted tokens, equipped with 3D coordinates from the \textbf{Head Decoder}, to many-to-one aggregation in $\mathcal{E}$ or one-to-one merging into $\mathcal{G}$; when $\mathcal{G}$ is full, it triggers re-merging to free a slot for the incoming merged token, then stores the tokens in a 3D voxel grid for future retrieval.}
\caption{\textbf{Overview of STAC.} Our framework reconstructs 3D scenes online using spatio-temporal token caching and chunk-based causal inference.
(a) The \textbf{Causal-VGGT} module processes ViT-tokenized frames in each chunk using causal attention over the working temporal cache $\mathcal M^{\text{temp}}$ and spatial cache $\mathcal M^{\text{spat}}$ retrieved from a 3D voxel grid. 
(b) During inference, \textbf{Working Temporal Token Caching} updates token scores after each KV cache access, retaining high-scoring anchor tokens $\mathcal M^{\text{anchor}}$ while preserving first-frame reference tokens $\mathcal M^{\text{refer}}$ and sliding-window tokens $\mathcal M^{\text{window}}$, and evicting the rest. 
(c) \textbf{Long-term Spatial Token Caching} routes evicted tokens with 3D coordinates from the \textbf{Head Decoder} to many-to-one aggregation in $\mathcal{E}$ or one-to-one merging into $\mathcal{G}$. When $\mathcal{G}$ is full, re-merging frees a slot for the incoming merged token, and the updated representations are stored in a 3D voxel grid for future retrieval.}
    \label{fig:framework}
\vspace{-10pt}
\end{figure*}

\section{Method}
\label{sec:methodology}

In this section, we introduce \textbf{STAC}, a training-free KV cache compression framework for causal transformer-based streaming 3D reconstruction, based on the observations in Sec.~\ref{sec:observation}. As shown in Fig.~\ref{fig:framework}, STAC consists of two complementary components: \textit{Working Temporal Token Caching} (Sec.~\ref{subsec:temp}), which integrates global reference, local window, and anchor tokens to maintain temporal coherence; and \textit{Long-term Spatial Token Caching} (Sec.~\ref{subsec:spatial}), which organizes evicted tokens in a 3D voxel grid and merges them online for efficient spatial reuse. In addition, \textit{Chunk-based Multi-frame Optimization} (Sec.~\ref{subsec:chunk}) jointly processes consecutive frames to improve both reconstruction accuracy and computational efficiency.

\subsection{Working Temporal Token Caching}
\label{subsec:temp}

The working temporal cache maintains a compact short-term memory over recent frames while selectively preserving a set of persistent tokens that remain important over longer spans. This design exploits the fact that, in streaming reconstruction, relevant context mainly comes from recent observations and a few globally informative cues. For simplicity, we omit layer and head indices.
Given a continuous stream of input frames, we construct the working temporal cache at time step $t$ by integrating three types of tokens:

(1) \textit{Global reference tokens} $\mathcal M^{\text{refer}}$: the tokens from the first frame $\mathbf{m}_1$ serve as a global reference, enabling all subsequent frames to be incrementally aligned within a unified coordinate system for stable and consistent reconstruction.

(2) \textit{Local window tokens} $\mathcal M_t^{\text{window}}$: a sliding temporal window of size $s$ captures short-term temporal correlations and motion continuity, using tokens from $\{\mathbf{m}_{t-s}, \ldots, \mathbf{m}_{t-1}\}$.

(3) \textit{Anchor tokens} $\mathcal{M}_t^{\text{anchor}}$: persistent informative tokens, including landmark-frame and camera tokens, encoding geometric or camera-related information and dynamically maintained to preserve long-range temporal context.

Accordingly, the working temporal cache at time step $t$ is defined as:
% \begin{equation}
% \mathcal M^{\text{temp}}_t = \mathbf{m}_1 \cup \{\mathbf{m}_{t-s}, \ldots, \mathbf{m}_{t-1}\} \cup \mathcal M_t^{\text{anchor}},
% \end{equation}
\begin{equation}
\mathcal M^{\text{temp}}_t = \mathcal M^{\text{refer}} \cup \mathcal M_t^{\text{window}} \cup \mathcal M_t^{\text{anchor}},
\end{equation}
While $\mathcal M^{\text{refer}}$ is fixed and $\mathcal M_t^{\text{window}}$ is updated by a sliding-window rule, $\mathcal M_t^{\text{anchor}}$ is dynamically maintained under a budget using attention-based importance scores. We next detail the score update and Top-$K$ anchor selection.

\textbf{Importance score update.}
To track persistent anchor tokens, we maintain a decaying importance score $s_i$ for each token in the temporal cache $\mathcal{M}_t^{\text{temp}}$. At time step $t$, the current frame produces query features $\{{q}_t^j\}_{j=1}^N$, and attention weights are computed using scaled dot-product attention:
\begin{equation}
\alpha_t^{j,i} = \mathrm{Softmax}_i\left({q}_t^j \cdot {\mathbf{{k}}}_t^{\text{all}} / {\sqrt{d_h}}\right),
\label{eq:attn}
\end{equation}
where the Softmax is taken over the keys ${\mathbf{k}}_t^{\text{all}}$ from the working temporal cache $\mathcal{M}_t^{\text{temp}}$, the spatial cache $\mathcal{M}_t^{\text{spat}}$ (Sec.~\ref{subsec:spatial}), and the current-frame tokens $\mathbf{m}_t$, for each query $q_t^j$. Here, $d_h$ denotes the per-head feature dimension. The importance scores are then updated with exponential decay:
{\begin{equation}
s_t^i = \gamma s_{t-1}^i + \sum_{j=1}^{N} \alpha_t^{j,i}, \quad  i \in \mathcal I(\mathcal{M}_t^{\text{temp}}),
\label{eq:scores}
\end{equation}
}
where $\gamma \in (0,1)$ is a decay factor and $\mathcal{I}(\cdot)$ denotes the index set of cached tokens. 
For newly generated tokens in the current frame, we initialize
$s_t^i = \textstyle\sum_{j}\alpha_t^{j,i}$.
%=============
% To track persistent anchor tokens, we maintain a decaying importance score $s_i$ for each token in the temporal cache $\mathbf{M}_t^{\text{temp}}$. At time step $t$, the current frame produces query features $\{{q}_t^j\}_{j=1}^N$, and attention weights are computed using scaled dot-product attention:
% \begin{equation}
% \alpha_t^{j,i} = \mathrm{Softmax}_i\left( \frac{{q}_t^j \cdot {k}_i}{\sqrt{d_h}} \right),
% \end{equation}
% where the Softmax is applied over all key indices $i$ from $\mathbf{M}_t^{\text{temp}}$, $\mathbf{M}_t^{\text{spatial}}$ (detailed in Sec.~\ref{subsec:spatial}), and $\mathbf{m}_t$ for each fixed $j$. Here, $k_i$ is a key vector retrieved from one of these caches, and $d_h$ denotes the token feature dimension.

% To accumulate evidence of token relevance over time, we update the importance score $s_t^i$ by summing the attention weights from all current-frame queries and applying exponential decay to prior values:
% \begin{equation}
% s_t^i = \gamma \cdot s_{t-1}^i + \sum_{j=1}^{N} \alpha_t^{j,i}, \quad \forall i \in \mathcal{I}(\mathbf{M}_t^{\text{temp}}),
% \label{eq:scores}
% \end{equation}
% where $\gamma \in (0,1)$ is a decay factor and $\mathcal{I}(\cdot)$ denotes the index set of cached tokens. For newly generated tokens in the current frame, their initial score $s_t^i$ is simply set to the sum of attention weights they receive: $s_t^i = \sum_{j=1}^N \alpha_t^{j,i}$.
%==========

\textbf{Anchor token selection.}  
After updating the importance scores, we retain the $\text{Top-K}$ most informative tokens as the new anchor tokens:
% \begin{equation}
% \mathcal M_{t+1}^{\text{anchor}} = \text{Top-K}(\{\mathcal m_i\}, \{s_i\}), \quad i \in \mathcal{I}(\mathcal M_t^{\text{anchor}}) \cup \mathcal{I}(\mathbf m_{t-s}),
% \end{equation}
\begin{equation}
\mathcal M_{t+1}^{\text{anchor}} = \text{Top-K}(\{\mathbf m_i\}, \{s_i\}), \quad i \in \mathcal I(\mathcal M_t^{\text{anchor}} \cup \mathbf m_{t-s}),
\end{equation}
where $\mathbf m_{t-s}$ represents the tokens that are about to be discarded from the sliding window. This selective update ensures that only salient and temporally stable features are persistently cached, enabling a compact yet expressive temporal memory representation.

\subsection{Long-term Spatial Token Caching}
\label{subsec:spatial}

While the working temporal cache captures local coherence, evicting tokens in long sequences discards early spatial evidence and harms later reconstruction. Since retaining all tokens is memory-prohibitive~\cite{zhuo2025streaming, lan2025stream3r}, we propose \textbf{Long-term Spatial Token Caching}, which exploits 3D spatial redundancy with a voxel-based structure (Sec.~\ref{sec:observation}). Tokens are assigned to voxels by 3D position and compressed independently. Each voxel maintains a dual-cache: a short-term buffer $\mathcal{E}_u$ for recently evicted tokens and a long-term set $\mathcal{G}_u$ of merged representatives, preserving feature diversity for long-term spatial reuse.

\textbf{Token Merging.}
To preserve information from evicted tokens under a fixed memory budget, we cast cache updates as a capacity-constrained online clustering problem. It comprises two complementary operations: \textit{one-to-one merging}, which absorbs an evicted token into its most similar long-term representative to reduce redundancy, and \textit{many-to-one aggregation}, which summarizes buffered tokens into a single representative inserted as a new cluster center to capture novel observations. Given an evicted token $m^e=\{k^e,v^e\}\in\mathbb{R}^{2\times d_h}$ assigned to voxel $u$, we update the cache as follows. Pseudocode and implementation details are provided in Sec.~B of the Supplementary Material.

\textit{Merging with Long-term Tokens.}  
To eliminate redundancy while reinforcing frequently observed spatial patterns, we match $m^e$ against representatives $\hat m\in\mathcal{G}_u$ using cosine similarity on key embeddings:
\begin{equation}
\hat m^p = \arg\max_{\hat m \in \mathcal{G}_u} \cos_k(m^e, \hat m),
\end{equation}
where $\cos_{k}(m^e, \hat m)$ is the cosine similarity in the key space. If $\cos_k(m^e, \hat m^p) > \lambda$, a weighted fusion is performed:
\begin{equation}
\hat m^p \leftarrow \frac{Z(\hat m^p) \hat m^p + \omega(m^e, \hat m^p) m^e}{Z(\hat m^p) + \omega(m^e, \hat m^p)}
\end{equation}
where $\omega(\cdot,\cdot)=\exp(\cos_k(\cdot,\cdot))$ following~\cite{wan2024d2o}. The cumulative weight is updated as $Z(\hat m^p) \leftarrow Z(\hat m^p) + \omega(m^e, \hat m^p)$. Otherwise, $m^e$ is enqueued into $\mathcal{E}_u$ to prevent prematurely merging potentially distinctive evidence; this also handles the case where $\mathcal{G}_u$ is initially empty.

\textit{Aggregation of Evicted Tokens.}
To preserve rare yet informative observations that fail the similarity threshold, we buffer dissimilar tokens in $\mathcal{E}_u$ and aggregate them once the buffer reaches capacity:
\begin{equation}
\hat{m} = \frac{\sum_{m \in \mathcal{E}_u} \omega(m, m^q) m}{\sum_{m \in \mathcal{E}_u} \omega(m, m^q)},
\end{equation}
where $m^q$ is the highest-scored (Eq.~\ref{eq:scores}) pivot token in buffer $\mathcal{E}_u$, as illustrated in Fig.~\ref{fig:framework}. The cumulative weight is assigned as $Z(\hat{m}) = \sum_{m \in \mathcal{E}_u} \omega(m, m^q)$, after which $\hat{m}$ is inserted into $\mathcal{G}_u$ as a new representative.

\textit{Re-merging under memory constraints.}
Finally, when \(\mathcal{G}_u\) reaches its capacity, we trigger a lightweight compaction step to make room for the incoming merged token.
We select the least informative representative \(\hat m^l=\arg\min_{\hat m\in\mathcal{G}_u} Z(\hat m)\), fuse it into its most similar neighbor in \(\mathcal{G}_u\setminus\{\hat m^l\}\), and then discard \(\hat m^l\) to free one slot; details are provided in Sec.~B.2 of the Supplementary Material.
% {
%     \begin{equation}
%     \begin{aligned}
%     &\hat{w}^{l} = e^{-1}Z(m^l) \omega (m^p,m^l), \\
%     &m^p \leftarrow \frac{Z(m^p)m^p + \hat{w}^{l}m^l}{Z(m^p) + \hat{w}^{l}},
%     \end{aligned}
%     \end{equation}
% }

Our design provides three key advantages: (1) {memory efficiency}, enabled by voxel-aware compression; (2) {long-term consistency}, achieved by retaining early-scene information; and (3) {feature diversity}, preserved via a dual-cache mechanism that maintains heterogeneous representations. Collectively, these properties support scalable and temporally consistent memory for streaming 3D reconstruction.

% %2025.02 End=============================================

\textbf{Token Retrieval.}
To provide spatial context during decoding, we retrieve relevant tokens based on the voxel grid. At each time step $t$, the model decodes per-pixel 3D points and voxelizes them to obtain the visible voxel set $\mathcal{V}_t$. We then expand it to $\mathrm{Nbr}(\mathcal{V}_t)$ by k-nearest neighbor (kNN) search over voxel centers to fetch nearby voxels, and retrieve all tokens from the long-term and temporary buffers:
\begin{equation}
\mathcal M^{\text{spat}}_{t}
= \left\{ m \,\middle|\, 
m \in \mathcal{G}_u\cup\mathcal{E}_u,\;
u \in \mathrm{Nbr}(\mathcal{V}_t)
\right\}.
\end{equation}
This localized retrieval ensures that only spatially and semantically relevant tokens are accessed, improving both efficiency and consistency.
To compensate for identity loss caused by token merging, we follow~\cite{tian2025keepkv, bolya2022token} and apply a \emph{count-based bias} to correct attention logits. Each merged token carries a count $n$ and modifies the attention as:
\begin{equation}
q\cdot k / \sqrt{d_h} + \log n,
\end{equation}
boosting frequently merged tokens during attention (Eq.~\ref{eq:attn}).

\subsection{Chunk-based Multi-frame Optimization}
\label{subsec:chunk}
To improve GPU utilization and reconstruction accuracy during streaming inference, we adopt a \emph{chunk-based multi-frame optimization} strategy. 
Rather than processing frames strictly one by one, we group a small set of consecutive available frames into temporal chunks and process them jointly. 
This procedure is \emph{chunk-causal}: at time step $t$, each chunk is constructed using only frames available up to $t$ (i.e., no frames with index $>t$ are used).
Within a chunk, attention is computed {bidirectionally} across frames, enabling limited intra-chunk information exchange to enhance local geometric consistency, while the chunk boundary maintains the streaming constraint. 
In practice, chunk-level execution amortizes kernel-launch overhead and data transfers, thereby substantially improving GPU efficiency compared to frame-by-frame processing.

For memory efficiency, we voxelize cached tokens and map 3D voxel coordinates to locality-preserving indices via Morton codes~\cite{morton1966computer, connor2010fast}. 
We further batch token selection, merging, and retrieval across layers after the forward pass to reduce KV cache overhead.

\begin{figure*}[htbp]
    \centering
    \includegraphics[width=0.95\linewidth]{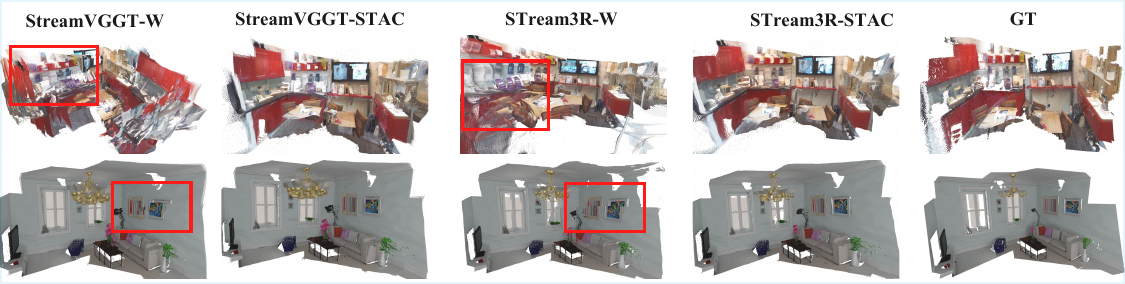}

    \caption{Qualitative results on {streaming } inputs from the 7-scenes and NRGBD datasets.}
    \label{fig:qulity}
    \vspace{-8pt}
\end{figure*}

\begin{table*}[t]
    \centering
% \caption{Quantitative results for point cloud reconstruction on the NRGBD and 7 Scenes datasets. Integrating STAC into STream3R and StreamVGGT substantially improves memory and runtime efficiency while preserving reconstruction quality. W8 denotes a sliding window size of 8. W24 and W30 indicate larger windows for fair total memory, though our method uses less memory at runtime. \textbf{Mem (GB):} reported as main value (including spatial token caching overhead) and parenthesized value (measured runtime usage).}
\caption{Quantitative results for point cloud reconstruction on the NRGBD and 7 Scenes datasets. Integrating STAC into STream3R and StreamVGGT substantially improves memory and runtime efficiency while preserving reconstruction quality. W8 denotes a sliding window size of 8. W22 and W26 indicate larger windows for fair total memory, though our method uses less memory at runtime. \textbf{Mem:} main values report total memory footprint for comparison; for STAC, this includes the spatial token cache storage, while parentheses show its runtime usage. \textbf{FPS:} end-to-end throughput measured over the full reconstruction pipeline, from image input to the final 3D outputs.}
    \resizebox{1.0\textwidth}{!}{%
    \begin{tabular}{l|c|cc|cc|cc|cc|cc|cc|c|c}
        \toprule
        \multirow{3}{*}{\textbf{Method}} & \multirow{3}{*}{\textbf{Type}} 
        & \multicolumn{6}{c|}{\textbf{NRGBD (stride 5)}} 
        & \multicolumn{6}{c|}{\textbf{7-Scenes (stride 5)}} 
        & \multirow{3}{*}{\textbf{Mem (GB)}~$\downarrow$} 
        & \multirow{3}{*}{\textbf{FPS~$\uparrow$}} \\
        % \cline{3-14}\midrule
        & & \multicolumn{2}{c|}{\textbf{ACC~$\downarrow$}} 
          & \multicolumn{2}{c|}{\textbf{Comp~$\downarrow$}} 
          & \multicolumn{2}{c|}{\textbf{NC~$\uparrow$}} 
          & \multicolumn{2}{c|}{\textbf{ACC~$\downarrow$}} 
          & \multicolumn{2}{c|}{\textbf{Comp~$\downarrow$}} 
          & \multicolumn{2}{c|}{\textbf{NC~$\uparrow$}} 
        \\
        & & mean & med & mean & med & mean & med & mean & med & mean & med & mean & med & & \\
        \midrule
        DUST3R-GA~\cite{wang2024dust3r} & Optim & 0.619 & 0.384 & 0.112 & 0.028 & 0.504 & 0.497 & 0.406 & 0.266 & 0.184 & 0.055 & 0.523 & 0.532 & -- & $<$1 \\
        MASt3R-GA~\cite{leroy2024grounding} & Optim & 0.599 & 0.311 & 0.306 & 0.096 & 0.552 & 0.570 & 0.459 & 0.266 & 0.187 & 0.040 & 0.544 & 0.568 & -- & $<$1 \\
        MonST3R-GA~\cite{zhang2024monst3r} & Optim & 0.528 & 0.399 & 0.242 & 0.098 & 0.550 & 0.576 & 0.328 & 0.214 & 0.200 & 0.031 & 0.529 & 0.545 & -- & $<$1 \\
        \midrule
        CUT3R~\cite{wang2025continuous} & Online & 0.219 & 0.142 & 0.098 & 0.036 & 0.612 & 0.686 & 0.097 & 0.050 & 0.042 & 0.010 & 0.585 & 0.632 & 0.18 & 19.79 \\
        Spann3R~\cite{wang2024spann3r} & Online & 0.068 & 0.037 & 0.020 & 0.006 & 0.637 & 0.733 & 0.052 & 0.020 & 0.019 & 0.006 & 0.584 & 0.628 & 0.37 & 57.21 \\
        Point3R~\cite{wu2025point3r} & Online & 0.095 & 0.047 & 0.021 & 0.005 & 0.668 & 0.794 & 0.049 & 0.023 & 0.033 & 0.013 & 0.595 & 0.649 & 1.50 & 8.34 \\
     \midrule
        VGGT~\cite{wang2025vggt} & Offline & {0.017} & {0.010} &
        {0.012} & {0.003} & {0.740} & {0.881} & {0.022} & {0.008} & 0.024 & 0.008 & 0.602 & 0.658 & -- & $<$1 \\

        \midrule
        STream3R~\cite{lan2025stream3r} & Online & \textbf{0.053} & \textbf{0.023} & \textbf{0.013} & \underline{0.005} & \textbf{0.703} & \textbf{0.850} & \textbf{0.044} & \textbf{0.013} & \textbf{0.024} & \underline{0.007} & \textbf{0.606} & \textbf{0.666} &  19.75  & 2.52 \\
        STream3R-W8 & Online & 0.078 & 0.036 & 0.015 & \underline{0.005} & 0.687 & 0.831 & 0.107 & 0.037 & 0.035 & 0.008 & 0.587 & 0.635 & \textbf{0.86} & \underline{6.19} \\
        STream3R-W22 & Online & 0.088 & 0.042 & 0.019 & 0.006 & 0.689 & 0.839 & 0.102 & \underline{0.029} & \underline{0.029} & \textbf{0.006} & \underline{0.594} & \underline{0.647} & \underline{2.19} & 5.24 \\
        STream3R-STAC & Online & \underline{0.065} & \underline{0.026} & \underline{0.014} & \textbf{0.004} & \underline{0.700} & \underline{0.847} & \underline{0.047} & \textbf{0.013} & \textbf{0.024} & \textbf{0.006} & \textbf{0.606} & \textbf{0.666} & {2.20}(\textbf{0.86}) & \textbf{10.53} \\
        \midrule
        StreamVGGT~\cite{zhuo2025streaming} & Online & \underline{0.134} & \underline{0.091} & \underline{0.059} & 0.012 & \underline{0.651} & \underline{0.772} & \textbf{0.046} & \textbf{0.026} & \textbf{0.027} & \textbf{0.007} & \underline{0.595} & \underline{0.648} &  19.75  & 2.48 \\
        StreamVGGT-W8 & Online & 0.168 & 0.101 & 0.065 & 0.016 & 0.647 & 0.767 & 0.136 & 0.058 & 0.042 & \textbf{0.007} & 0.584 & 0.631 & \textbf{0.86} & \underline{6.10} \\
        StreamVGGT-W26 & Online & 0.174 & 0.121 & 0.061 & \underline{0.011} & 0.634 & 0.740 & 0.117 & 0.058 & 0.050 & \underline{0.010} & 0.579 & 0.623 & \underline{2.57} & 5.41 \\
        StreamVGGT-STAC & Online & \textbf{0.126} & \textbf{0.081} & \textbf{0.047} & \textbf{0.010} & \textbf{0.682} & \textbf{0.829} & \underline{0.056} & \underline{0.030} & \underline{0.029} & \textbf{0.007} & \textbf{0.596} & \textbf{0.650} & 2.57(\textbf{0.86}) & \textbf{10.49} \\
        \bottomrule
     
    \end{tabular}
    }
    \label{tab:3drecon}
    \vspace{-8pt}
\end{table*}

\begin{table*}[htbp]
  \centering
  \caption{Quantitative camera pose estimation results on the Sintel, TUM Dynamics, and ScanNet datasets.}
  \resizebox{0.95\textwidth}{!}{
  \begin{tabular}{l|c|ccc|ccc|ccc}
    \toprule
    \multirow{2}{*}{\textbf{Method}} & \multirow{2}{*}{\textbf{Type}} 
    & \multicolumn{3}{c|}{\textbf{Sintel}} 
    & \multicolumn{3}{c|}{\textbf{TUM}} 
    & \multicolumn{3}{c}{\textbf{ScanNet}} \\
    & & \textbf{ATE$\downarrow$} & \textbf{RPE trans$\downarrow$} & \textbf{RPE rot$\downarrow$} 
      & \textbf{ATE$\downarrow$} & \textbf{RPE trans$\downarrow$} & \textbf{RPE rot$\downarrow$} 
      & \textbf{ATE$\downarrow$} & \textbf{RPE trans$\downarrow$} & \textbf{RPE rot$\downarrow$} \\
    \midrule
    DUST3R-GA~\cite{wang2024dust3r} & Optim & 0.417 & 0.250 & 5.796 & 0.083 & 0.017 & 3.567 & 0.081 & 0.028 & 0.784 \\
    MASt3R-GA~\cite{leroy2024grounding} & Optim & 0.185 & 0.060 & 1.496 & 0.038 & 0.012 & 0.448 & 0.078 & 0.020 & 0.475 \\
    MonST3R-GA~\cite{zhang2024monst3r} & Optim & 0.111 & 0.044 & 0.869 & 0.098 & 0.019 & 0.935 & 0.077 & 0.018 & 0.529 \\
    \midrule
    CUT3R~\cite{wang2025continuous} & Online & 0.213 & 0.066 & 0.621 & 0.046 & 0.015 & 0.473 & 0.099 & 0.022 & 0.600 \\
    Spann3R~\cite{wang2024spann3r} & Online & 0.329 & 0.110 & 4.471 & 0.056 & 0.021 & 0.591 & 0.096 & 0.023 & 0.661 \\
    Point3R~\cite{wu2025point3r} & Online & 0.351 & 0.128 & 1.822 & 0.075 & 0.029 & 0.642 & 0.106 & 0.035 & 1.946 \\
    \midrule
    VGGT~\cite{wang2025vggt} & Offline & 0.174 & 0.056 & 0.465 & 0.013 & 0.010 & 0.311 & 0.036 & 0.015 & 0.377 \\
    \midrule
    STream3R~\cite{lan2025stream3r} & Online & \underline{0.219} & \textbf{0.065} & \textbf{0.867} & \textbf{0.026} & \textbf{0.013} & \textbf{0.331} & \textbf{0.052} & \textbf{0.021} & \textbf{0.850} \\
    STream3R-W8 & Online & 0.359 & \underline{0.081} & \underline{1.160} & 0.032 & 0.015 & \underline{0.351} & 0.115 & 0.040 & 2.223 \\
    STream3R-STAC & Online & \textbf{0.198} & 0.082 & 1.486 & \underline{0.030} & \underline{0.014} & \underline{0.351} & \underline{0.053} & \underline{0.025} & \underline{0.983} \\
    \midrule
    StreamVGGT~\cite{zhuo2025streaming} & Online & \textbf{0.224} & \underline{0.092} & \textbf{0.721} & \textbf{0.026} & \textbf{0.012} & \textbf{0.314} & \textbf{0.048} & \textbf{0.019} & \textbf{0.477} \\
    StreamVGGT-W8 & Online & 0.409 & \textbf{0.087} & \underline{0.918} & \underline{0.043} & 0.015 & 0.331 & 0.112 & 0.030 & 1.459 \\
    StreamVGGT-STAC & Online & \underline{0.251} & 0.102 & 1.341 & \textbf{0.026} & \underline{0.013} & \underline{0.326} & \underline{0.059} & \underline{0.022} & \underline{0.531} \\
    \bottomrule
  \end{tabular}
  }
  \label{tab:cam}
  \vspace{-8pt}
\end{table*}

\section{Experiments}

\subsection{Implementation Details}
We extend Causal-VGGT~\cite{lan2025stream3r, zhuo2025streaming} to support memory-constrained online 3D reconstruction and camera pose estimation. We set the decay factor to $\gamma=0.9$, voxel grid resolution to $0.05$, and cosine similarity threshold for token merging to $\lambda=0.8$. Each voxel stores at most $|\mathcal{G}|=4$ merged tokens and $|\mathcal{E}|=8$ temporarily evicted tokens, with kNN retrieval performed within a $2\times$ voxel radius.
% To enable merge-aware attention, we implement a Triton~\cite{tillet2019triton} operator that computes attention output and importance scores (Eq.~\ref{eq:scores}) over a subsampled set of queries for efficiency.
To enable merge-aware attention, we implement a custom CUDA kernel that computes attention outputs and importance scores.
At runtime, the first frame’s KV cache is fully retained. The remaining budget, set to $8\times$ the number of frame tokens, is allocated as follows: 50\% to the temporal window ($s=4$), 25\% to anchor tokens, and 25\% to retrieved merged tokens (with evicted tokens used when insufficient). All KV caches are stored in \texttt{float16} to reduce memory consumption. We use chunk size of 4 for multi-frame optimization, and all experiments run on a single 40GB A100 GPU.

\textbf{Baselines.}
Thanks to its plug and play design, STAC can be seamlessly integrated into Causal-VGGT style causal transformers, including STream3R~\cite{lan2025stream3r} and StreamVGGT~\cite{zhuo2025streaming}. We also compare sliding window variants of STream3R and StreamVGGT, namely STream3R-W and StreamVGGT-W, which retain the first frame and apply a fixed length temporal window during causal inference, following prior practice~\cite{zhang2023h2o,xiao2023efficient,lan2025stream3r}. For broader comparison, we evaluate online baselines including Spann3R~\cite{wang2024spann3r}, CUT3R~\cite{wang2025continuous}, and Point3R~\cite{wu2025point3r}, which incorporate memory mechanisms but are tightly coupled with architectures in the DUSt3R family and typically require fine tuning. In addition, we consider pair based methods such as DUSt3R~\cite{wang2024dust3r}, MASt3R~\cite{leroy2024grounding}, and MonST3R~\cite{zhang2024monst3r}, which rely on global alignment modules to process streaming inputs.

\subsection{Evaluation}

\textbf{3D Reconstruction.}
We evaluate STAC on the NRGBD~\cite{azinovic2022neural} and 7-Scenes~\cite{shotton2013scene} datasets, using input images of resolution $518\times392$. For each sequence, we sample keyframes every 5 frames, resulting in 200–300 frames per clip, approximating streaming scenarios. Following prior works~\cite{wang2024dust3r, wang2024spann3r, wu2025point3r}, we report Accuracy (Acc), Completion (Comp), and Normal Consistency (NC) as geometric metrics. For memory and runtime profiling, we measure total memory usage and end-to-end FPS over entire sequences.

Table~\ref{tab:3drecon} summarizes the quantitative results; trends in Fig.~\ref{fig:teaser}. Integrating STAC into STream3R and StreamVGGT leads to substantial reductions in memory usage while improving FPS. Meanwhile, reconstruction accuracy remains comparable to, or even exceeds, that of the original models, validating the benefit of explicitly modeling spatio-temporal sparsity within the cache. To further assess the effectiveness of our cache design, we compare STAC with STream3R-W and StreamVGGT-W. For fairness, we enlarge their sliding-window sizes so that all methods operate under the same memory budget. As shown in Table~\ref{tab:3drecon}, STAC consistently achieves higher reconstruction accuracy and lower computational overhead under identical memory constraints, demonstrating the scalability and efficiency of our spatio-temporal cache. 
%Note that the memory values in parentheses report the measured runtime usage, while the main values additionally include the overhead from spatial token caching.

Our method provides superior long-sequence reconstruction quality. While Spann3R~\cite{wang2024spann3r} and CUT3R~\cite{wang2025continuous} rely on implicit or latent memory and thus suffer from long-term drift or forgetting, and Point3R~\cite{wu2025point3r} incorporates spatial memory without explicit temporal reasoning, our dual-cache architecture jointly captures long-range temporal cues and structured spatial context. This yields a compact yet expressive memory representation that maintains stability and consistency throughout extended sequences. Quantitative visual comparisons are provided in Fig.~\ref{fig:qulity}, further illustrating the advantages of STAC.

% \textbf{Camera Pose Estimation.}
% Following \cite{wang2025continuous,lan2025stream3r}, we evaluate camera pose estimation on the Sintel~\cite{butler2012naturalistic}, TUM Dynamics~\cite{sturm2012benchmark}, and ScanNet~\cite{dai2017scannet} datasets. We report Absolute Trajectory Error (ATE), Relative Pose Error (RPE) in translation and rotation, all computed after Sim(3) Umeyama alignment with ground-truth trajectories~\cite{zhang2024monst3r,zhuo2025streaming}.

% As shown in Table~\ref{tab:cam}, our method achieves competitive performance compared to both online and optimization-based baselines. The frame-by-frame variant matches or surpasses STream3R on most metrics, achieves comparable results to CUT3R\cite{wang2025continuous} on Sintel, and outperforms it on TUM and ScanNet—demonstrating the robustness of our approach even in dynamic scenes. Notably, the frame-based variant consistently outperforms the chunk-based counterpart in pose accuracy. We hypothesize that this gap arises from a mismatch between the chunked inference strategy and the causal attention structure used during training, which may introduce temporal inconsistencies across frames.

\textbf{Camera Pose Estimation.}
Following~\cite{wang2025continuous, lan2025stream3r}, we evaluate the task of camera pose estimation on the Sintel~\cite{butler2012naturalistic}, TUM Dynamics~\cite{sturm2012benchmark}, and ScanNet~\cite{dai2017scannet} datasets. We report Absolute Trajectory Error (ATE) and Relative Pose Error (RPE) in both translation and rotation, computed after Sim(3) Umeyama alignment with ground truth trajectories~\cite{zhang2024monst3r, zhuo2025streaming}. As shown in Table~\ref{tab:cam}, STream3R-STAC and StreamVGGT-STAC achieve competitive pose accuracy compared to STream3R and StreamVGGT, while operating under a substantially reduced memory budget. The improvements observed on Sintel and TUM further highlight the robustness of our cache design in dynamic scenes, where maintaining long term temporal consistency presents a significant challenge.

\begin{table}[t]
    \centering
    \footnotesize
    \renewcommand{\arraystretch}{0.95}
    \caption{Ablation study of anchor tokens (AC), spatial token caching (SC), count-based bias (CB), and chunk-based optimization (CO), reporting reconstruction metrics, memory, and backbone runtime (excluding the image encoder and head decoders).}
    \resizebox{\columnwidth}{!}{
        \begin{tabular}{l|ccc|cc}
        \toprule
        \textbf{Method} &
        \textbf{Acc~$\downarrow$} & \textbf{Comp~$\downarrow$} & \textbf{NC~$\uparrow$} &
        \textbf{Mem (GB)} & \textbf{Runtime (ms)} \\
        \midrule
        Baseline & 0.0776 & 0.0150 & 0.6865 & 0.858 & 92.56 \\
        w/o AC   & 0.0725 & 0.0209 & 0.6991 & 1.901 & 61.36 \\
        w/o SC   & 0.0713 & 0.0199 & 0.6939 & \textbf{0.572} & \textbf{39.08} \\
        w/o CB   & 0.0666 & 0.0175 & 0.6973 & 2.063 & 56.08 \\
        w/o CO   & 0.0673 & 0.0156 & 0.6948 & 1.805 & 138.12 \\
        Full     & \textbf{0.0648} & \textbf{0.0142} & \textbf{0.6995} & 2.210 & 71.18 \\
        \bottomrule
        \end{tabular}
    }
    \label{tab:ablation}
    \vspace{-10pt}
\end{table}

% \subsection{Ablation Study}
% We conduct an ablation study on the NRGBD~\cite{azinovic2022neural} dataset. Specifically, we remove each key component in isolation, including Anchor Tokens, Count-based Bias, Spatial Token Caching, and Chunk-based Optimization. We use STream3R-W8~\cite{lan2025stream3r} as the baseline. Table~\ref{tab:ablation} summarizes reconstruction quality, memory usage, and runtime performance.

% % \textit{Anchor Cache (w/o AC).} Removing anchor caching tokens degrades temporal stability because persistent informative tokens encode critical geometric or camera related information.
% \textit{Anchor Cache (w/o AC).} Removing anchor caching tokens degrades temporal stability, as persistent informative tokens encode critical geometric or camera-related information.

% \textit{Spatial Cache (w/o SC).} Disabling spatial caching tokens reduces memory consumption and increases speed, but it leads to a clear degradation in reconstruction quality.

% \textit{Count-based Bias (w/o CB).} Removing Count-based Bias slightly degrades reconstruction quality, as it compensates for identity loss introduced by token merging.

% \textit{Chunk-based Optimization (w/o CO).} Without Chunk based Optimization, both reconstruction quality and runtime degrade, indicating that this component is essential for improving accuracy while maintaining efficient parallelism.
\subsection{Ablation Study}
We conduct an ablation study on the NRGBD~\cite{azinovic2022neural} dataset by removing each key component in isolation, including Anchor Tokens, Spatial Token Caching, Count-based Bias, and Chunk-based Optimization. We use STream3R-W8~\cite{lan2025stream3r} as the baseline. Table~\ref{tab:ablation} summarizes reconstruction quality, memory usage, and runtime performance.
\begin{itemize}\setlength{\itemsep}{0pt}
  \item \textit{Anchor Cache (w/o AC).} Removing anchor tokens degrades temporal stability due to the loss of persistent geometric and camera-related information.
  \item \textit{Spatial Cache (w/o SC).} Disabling spatial caching reduces memory usage and improves speed, but causes a clear drop in reconstruction quality.
  \item \textit{Count-based Bias (w/o CB).} Removing CB slightly degrades reconstruction quality, as it fails to compensate for identity loss from token merging.
  \item \textit{Chunk-based Optimization (w/o CO).} Without Chunk based Optimization, both quality and runtime degrade, indicating that this component is essential for improving accuracy while maintaining efficient parallelism.
\end{itemize}

\section{Conclusion}
We present STAC, a training-free framework for streaming 3D reconstruction with a spatio-temporal aware cache compression module. By modeling structured spatial and temporal sparsity in the key-value cache, STAC enables compact memory usage and temporally consistent reconstruction. The proposed working temporal token caching preserves informative tokens via decayed cumulative attention, while long-term spatial token caching compresses redundant features into voxel-based representations for efficient reuse. The chunk-based multi-frame optimization jointly refines consecutive frames, improving temporal coherence and exploiting GPU parallelism for real-time inference. Extensive experiments show that STAC achieves state-of-the-art reconstruction quality with substantially lower memory and computational cost compared to prior streaming transformers. Beyond performance improvements, the framework establishes a unified paradigm for causal 3D perception under constrained memory. Future work explores adaptive cache learning and multimodal extensions to further improve scalability and generalization.

\textbf{Limitation.}
STAC still has two primary limitations. First, the voxel-based spatial caching depends on a fixed resolution grid. In large or unbounded outdoor scenes, the number of active voxels increases with scene extent, leading to increased memory usage. A practical extension is to offload infrequently accessed tokens to external storage such as the CPU to reduce GPU memory pressure. Second, in highly dynamic environments, fast object motion can introduce inconsistent token representations, which affects the stability of the cache.

\section*{Acknowledgments}
This work was supported by the National Natural Science Foundation of China (U25A20444, 62025207) and the Fundamental and Interdisciplinary Disciplines Breakthrough Plan of the Ministry of Education of China (JYB2025XDXM113). 

{
    \small
    \bibliographystyle{ieeenat_fullname}
    \bibliography{main}

}

% % % % WARNING: do not forget to delete the supplementary pages from your submission 
\clearpage
\appendix
\setcounter{page}{1}
% \clearpage
This supplementary material provides additional empirical observations, technical details, and extended experiments that complement the main paper.

\section{KV Cache Similarity}
\label{similarity}

\begin{figure}[t]
    \centering
    \includegraphics[width=\linewidth]{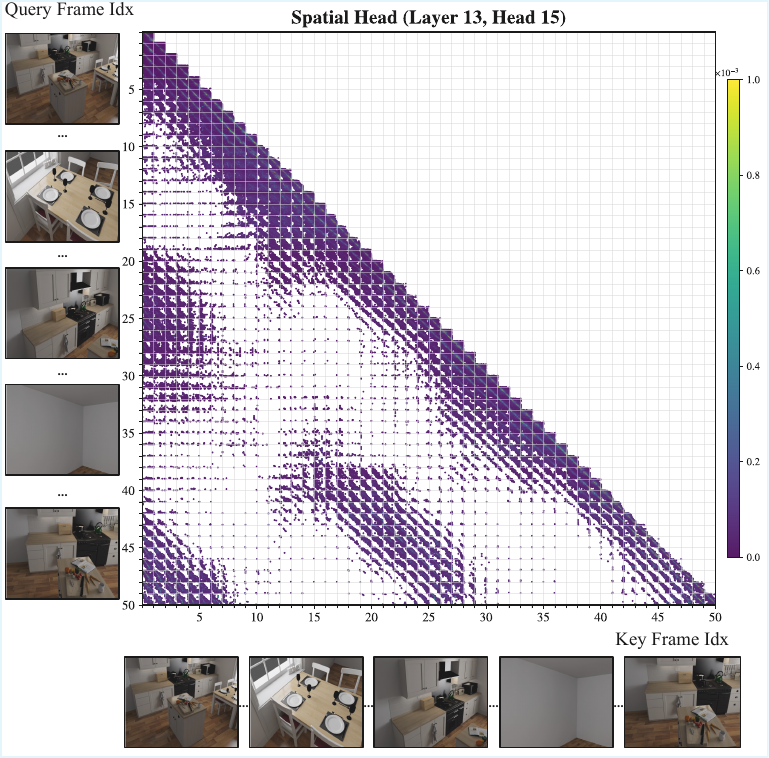}
    \caption{Visualization of a spatial attention head, where the top 1024 relevant key tokens are retained for each query. The axes are labeled at the frame level to facilitate identification of the corresponding frames.}
    \label{fig:spatial}
\end{figure}

% As streaming inputs accumulate over time, the Causal-VGGT model mitigates out-of-memory (OOM) issues by adopting a windowed attention strategy that retains only the KV caches of recent frames. However, this inevitably results in the loss of potentially important information contained in older KV tokens.

Building on the observation that spatial attention heads in Causal-VGGT~\cite{lan2025stream3r,zhuo2025streaming} are highly sensitive to viewpoint changes (Fig.~\ref{fig:spatial}), we investigate the spatial properties of their key/value (KV) tokens. In the context of streaming 3D reconstruction, the camera frequently revisits previously observed regions. When cache management relies solely on temporal correlations across viewpoints, relevant tokens from earlier views may be prematurely evicted due to viewpoint shifts. This motivates a deeper analysis of the spatial structure and similarity present in the KV cache.

We hypothesize that tokens in Causal-VGGT are aligned with the 3D scene structure, exhibiting both local redundancy\footnote{In our context, \emph{local similarity} and \emph{redundancy} refer to the same phenomenon of high feature similarity among spatially proximal tokens.} and distance-dependent similarity. To validate this hypothesis, we associate each key/value pair \(m_i=\{k_i, v_i\}\) with a 3D coordinate in the reconstructed scene and discretize these tokens into a voxel grid with fixed spatial resolution.  We then analyze: (i) \emph{intra-voxel similarity} (Fig.~\ref{fig:intra}), quantifying redundancy among tokens within the same spatial region; (ii) \emph{inter-voxel similarity} (Fig.~\ref{fig:inter}), characterizing how similarity decays with spatial separation; and (iii) \textit{layer-wise similarity} (Fig.~\ref{fig:layer}), measuring how consistently local redundancy is preserved across layers. Together, these analyses show that KV cache is locally redundant and spatially structured across Causal-VGGT's transformer layers, directly motivating the voxel-wise cache compression used in the \textbf{STAC} framework.
\subsection{Intra-voxel Similarity}
We first assess intra-voxel similarity to quantify local redundancy in the KV cache and understand how token representations cluster within the same spatial region.

For each voxel \(u \in \mathcal{V}\), we collect the set of $T_u$ tokens:
\begin{equation}
\mathcal{M}_u = \{m_{u,1}, m_{u,2}, \dots, m_{u,T_u}\}
\end{equation}
where each \(m_{u,i}\) is a token assigned to voxel \(u\).

We compute the pairwise cosine similarity between tokens as follows:
\begin{equation}
\begin{aligned}
\cos_k(m_i, m_j) &= \frac{k_i \cdot k_j}{\|k_i\| \|k_j\|},\\
\cos_v(m_i, m_j) &= \frac{v_i \cdot v_j}{\|v_i\| \|v_j\|}.
\end{aligned}
\label{eq:cos}
\end{equation}

To capture the internal structure within a voxel, we perform K-means clustering~\cite{Jin2010kmeans} over the key-state tokens in \(\mathcal{M}_u\), partitioning them into $n$ clusters:
\begin{equation}
\mathrm{Cluster}_u = \{ \mathcal{C}_{u,1}, \mathcal{C}_{u,2}, \dots, \mathcal{C}_{u,n} \},
\quad
\mathcal{C}_{u,i} \subseteq \mathcal{M}_u
\end{equation}

The intra-voxel similarity score is defined as the average cosine similarity between each token and its corresponding cluster centroid within a voxel:
\begin{equation}
\begin{aligned}
\mathrm{IntraSim}_k(u;n)
&=
\frac{1}{N_u(n)}
\sum_{i=1}^{n}
\sum_{\substack{m \in \mathcal{C}_{u,i}}}
\cos_k(\hat m_{u,i}, m),\\
\mathrm{IntraSim}_v(u;n)
&=
\frac{1}{N_u(n)}
\sum_{i=1}^{n}
\sum_{\substack{m \in \mathcal{C}_{u,i}}}
\cos_v(\hat m_{u,i}, m),
\end{aligned}
\end{equation}
where \(\hat m_{u,i} = \mathrm{Mean}(\mathcal{C}_{u,i})\) is the centroid of cluster \(\mathcal{C}_{u,i}\), and
\begin{equation}
N_u(n)=\sum_{i=1}^{n} |\mathcal{C}_{u,i}|
\end{equation}
is the total number of tokens across all clusters in voxel \(u\).

\begin{figure}[t]
    \centering
    \includegraphics[width=\linewidth]{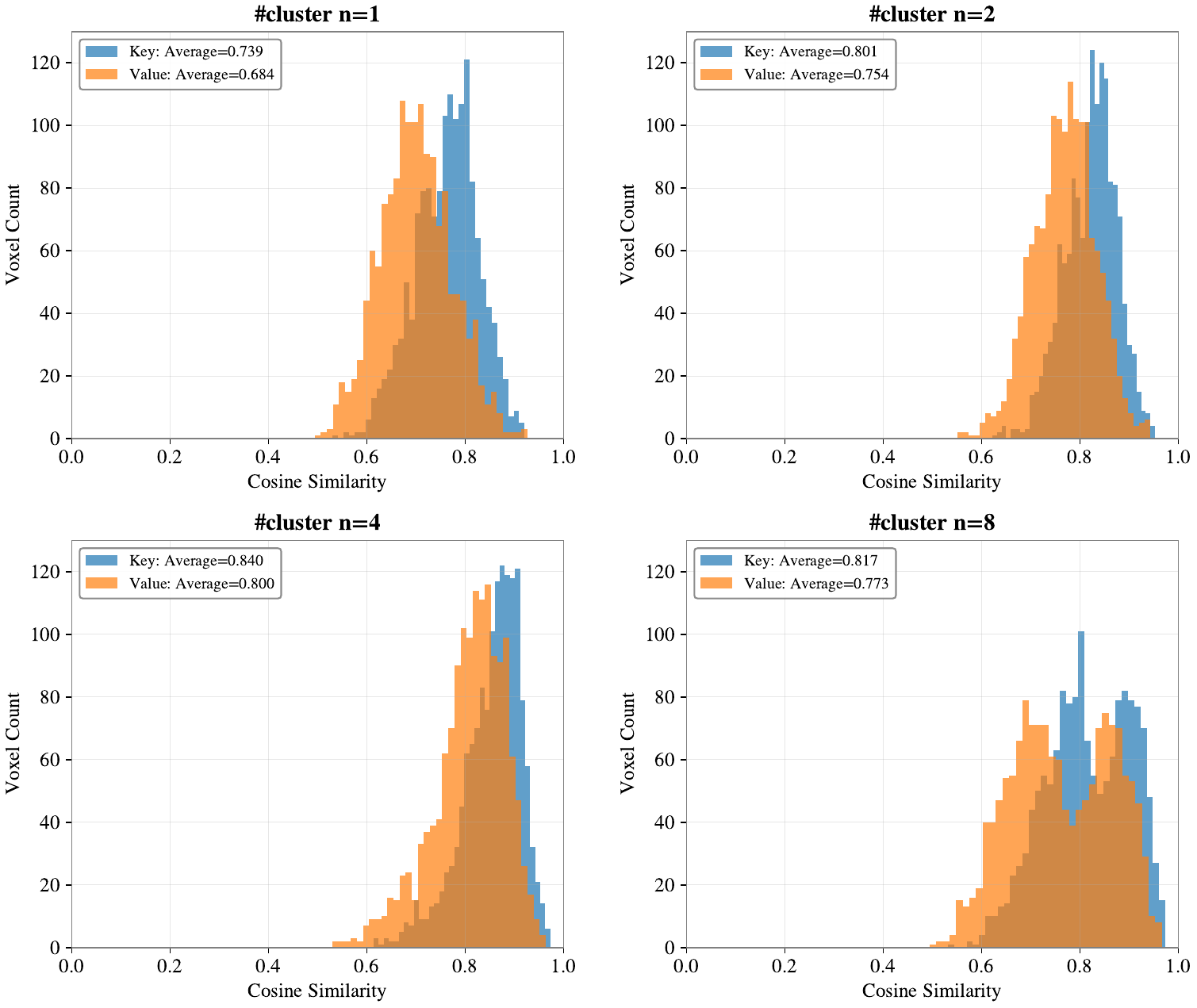}
    \caption{\textbf{Intra-voxel similarity distribution vs. cluster count.}
    Distribution of intra-voxel similarity scores across voxels under varying cluster counts \(n \in \{1,2,4,8\}\). Each score reflects the average cosine similarity of key/value tokens within voxel clusters.}
    \label{fig:intra}
    \vspace{-10pt}
\end{figure}

To evaluate the effect of clustering granularity, we compute intra-voxel similarity distributions across all voxels in a scene under varying cluster counts, as shown in Fig.~\ref{fig:intra}. The results reveal that key/value tokens within the same voxel generally exhibit strong mutual similarity, suggesting notable local redundancy in the KV cache. As the number of clusters $n$ increases, intra-cluster similarity improves due to finer partitioning. However, beyond $n=4$, the gains diminish, and smaller cluster sizes introduce noise due to insufficient token samples. Empirically, we find that $n=4$ provides a good trade-off between representational granularity and robustness. Notably, value-state tokens follow a similar trend, showing consistent behavior with their corresponding key tokens across different clustering levels.

\subsection{Inter-voxel Similarity}
\begin{figure}[t]
    \centering
    \includegraphics[width=\linewidth]{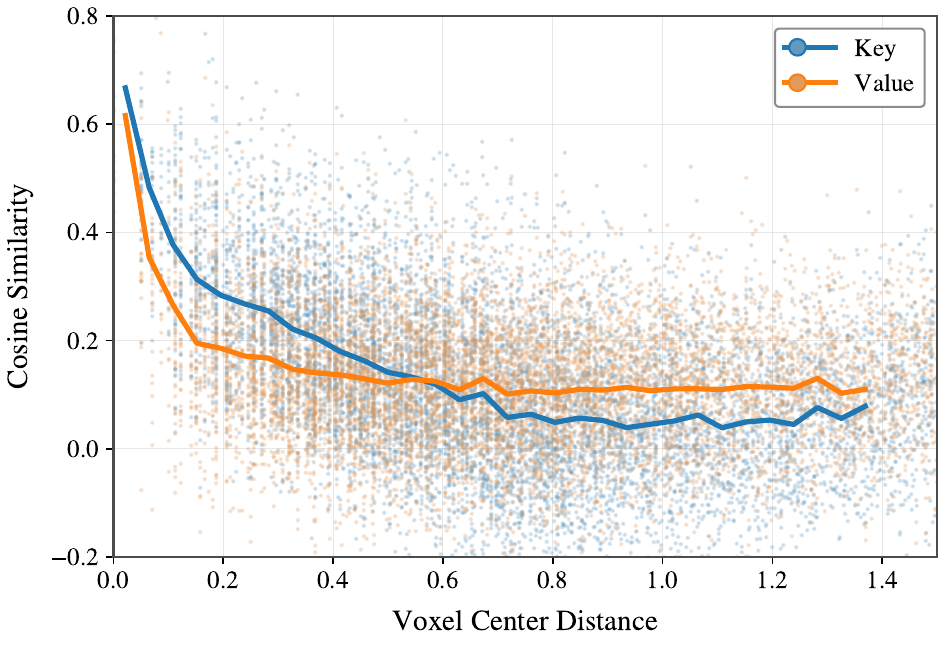}
    \caption{\textbf{Inter-voxel similarity vs. 3D distance.}
    Mean and samples of \(\mathrm{InterSim}(u_a, u_b)\) are shown as a function of Euclidean distance between voxel centers, averaged over sampled voxel pairs across scenes.}
    \label{fig:inter}
\end{figure}

We next evaluate inter-voxel similarity to characterize how token representations evolve across spatially distinct regions. By measuring feature similarity between tokens in different voxels, we aim to uncover the degree of spatial locality encoded in the KV cache and assess how it correlates with 3D distance.

Given two voxels \(u_a\) and \(u_b\) with associated token sets \(\mathcal{M}_{u_a}\) and \(\mathcal{M}_{u_b}\), we define their inter-voxel similarity in the key and value embedding spaces as follows:
\begin{equation}
\begin{aligned}
\mathrm{InterSim}_k(u_a, u_b)
&=
\frac{1}{|\mathcal{M}_{u_a}|\,|\mathcal{M}_{u_b}|}
\sum_{\substack{m_a \in \mathcal{M}_{u_a} \\ m_b \in \mathcal{M}_{u_b}}}
\cos_k(m_a, m_b),\\
\mathrm{InterSim}_v(u_a, u_b)
&=
\frac{1}{|\mathcal{M}_{u_a}|\,|\mathcal{M}_{u_b}|}
\sum_{\substack{m_a \in \mathcal{M}_{u_a} \\ m_b \in \mathcal{M}_{u_b}}}
\cos_v(m_a, m_b).
\end{aligned}
\end{equation}

To analyze spatial trends, we sample voxel pairs from the 3D grid, compute their inter-voxel similarities, and measure the Euclidean distance between voxel centers. As shown in Fig.~\ref{fig:inter}, similarity consistently decays with increasing distance: nearby voxels share highly similar key/value tokens, while distant ones become nearly orthogonal in feature space. This confirms that Causal-VGGT preserves spatial locality in its KV cache, yielding geometrically structured and semantically coherent representations.

Such findings reinforce the intuition behind our voxel-wise caching and compression scheme: restricting attention and aggregation to local 3D neighborhoods not only aligns with the spatial correlation of features but also mitigates the risk of conflating semantically unrelated content. This spatially localized organization thus serves as a principled foundation for efficient and effective representation compression in voxel space.

\subsection{Layer-wise Similarity}
\begin{figure}[t]
\centering
\includegraphics[width=\linewidth]{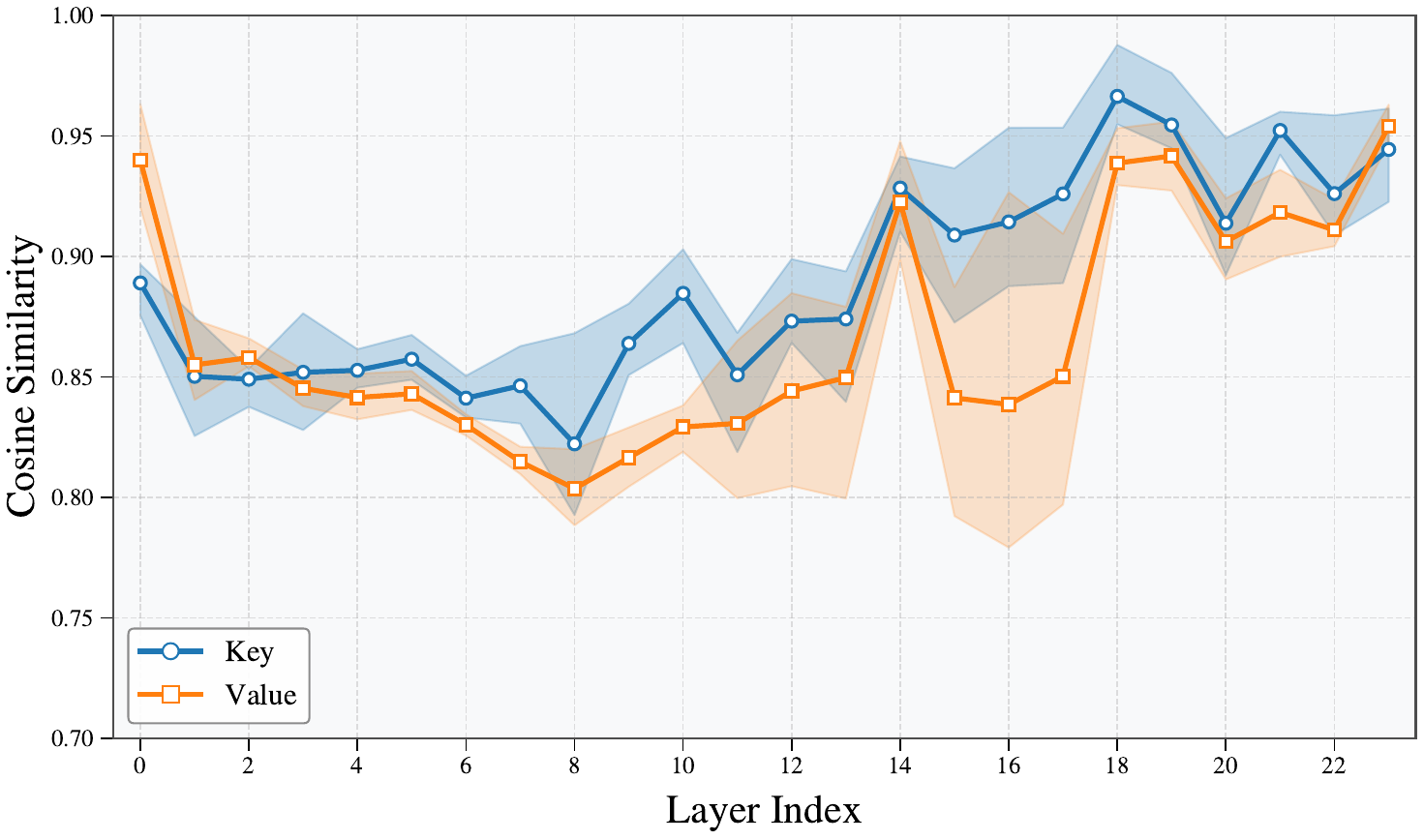}
\caption{\textbf{Layer-wise intra-voxel similarity in Causal-VGGT.}
Average intra-voxel similarity scores computed across all attention heads per layer.}
\label{fig:layer}
\end{figure}

We further analyze the intra-voxel similarity across different layers of the Causal-VGGT model. For each layer, we compute the similarity score per attention head and average them, as shown in Fig.~\ref{fig:layer}. The results indicate that Causal-VGGT consistently exhibits high intra-voxel similarity across all layers. We attribute this behavior to the model being trained specifically for 3D reconstruction tasks, which encourages the emergence of spatially coherent features. As a result, attention heads tend to capture features aligned with the same 3D regions across different views, leading to strong local redundancy in the key/value tokens at all layers. Given this consistency, our \textbf{STAC} framework adopts a unified cache compression strategy across all layers, without introducing layer- or head-specific selection heuristics.

\section{Details of Token Merging}

\subsection{Online Token Merging Algorithm}

Algorithm~\ref{alg:merge} summarizes the online token merging procedure within a local voxel \(u\). The goal is to maintain a compact merged token set \(\mathcal{G}_u\) in an online streaming setting with limited memory.
\begin{algorithm}[t]
\caption{Online Token Merging}
\label{alg:merge}
\begin{algorithmic}[1]
\Require Incoming tokens $\mathcal{M}_u$, merged buffer $\mathcal{G}_u$, evicted buffer $\mathcal{E}_u$, similarity threshold $\lambda$
\ForAll{$m \in \mathcal{M}_u$}
    \If{$\mathcal{G}_u \neq \emptyset$}
        \State $\hat m^p \gets \arg\max_{\hat m \in \mathcal{G}_u} \cos_k(m, \hat m)$
        \If{$\cos_k(m, \hat m^p) \geq \lambda$} \Comment{Merging}
            \State $w \gets \text{Weight}(\hat m^p, m)$
            \State $\mathcal{G}_u.\text{merge}(\hat m^p, m, w)$
        \Else
            \State $\mathcal{E}_u.\text{append}(m)$
        \EndIf
    \Else
        \State $\mathcal{E}_u.\text{append}(m)$
    \EndIf
    \If{$|\mathcal{E}_u| = N_{\text{evict}}$} \Comment{Aggregation}
        \State $\hat{m}_{new}, Z(\hat{m}_{new}) \gets \text{Aggregate}(\mathcal{E}_u)$ 
        \State $\mathcal{E}_u.\text{clear}()$
        \If{$|\mathcal{G}_u| = N_{\text{merge}}$} \Comment{Re-merging}
            \State $\hat m^l \gets \arg\min_{\hat m \in \mathcal{G}_u} Z(\hat m)$
            \State $\hat m^p \gets \arg\max_{\hat m \in \mathcal{G}_u \setminus \{\hat m^l\}} \cos_k(\hat m^l, \hat m)$
            \State $\hat{w}^l \gets e^{-1} Z(\hat m^l) \text{Weight}(\hat m^p, \hat m^l)$
            \State $\mathcal{G}_u.\text{merge}(\hat m^p, \hat m^l, \hat{w}^l)$
            \State $\mathcal{G}_u.\text{pop}(\hat m^l)$
        \EndIf
        \State $\mathcal{G}_u.\text{append}(\hat{m}_{new}, Z(\hat{m}_{new}))$
    \EndIf
\EndFor
\end{algorithmic}
\end{algorithm}
\subsection{Implementation Details for Token Merging}

%+++++++V3
In our merging module, each cluster \(\mathcal{C}_i\) is aggregated around a pivot token \(m_i^q\), selected according to its attention score. For notational simplicity, we omit the voxel index \(u\) in the following derivations, since all operations are performed within a fixed local voxel. We adopt a similarity-weighted aggregation scheme, following recent KV compression methods such as D2O~\cite{wan2024d2o}, so that tokens more similar to the pivot receive larger weights.
Formally, the merged token \(\hat{m}_i\) is computed as:
\begin{align}
\hat{m}_i &= \frac{1}{Z_i} \sum_{m \in \mathcal{C}_i} \omega(m_i^q, m)\, m,
 \\
\omega(m_i^q, m) &= \exp\left( \cos_k(m_i^q, m) \right), \label{eq:weight}\\
Z_i &= \sum_{m \in \mathcal{C}_i} \omega(m_i^q, m), \\
m_i^q &= \arg\max_{m \in \mathcal{C}_i} \mathrm{score}(m),
\end{align}
where \(\omega(\cdot,\cdot)\) is a cosine-based similarity weight (see Eq.~\eqref{eq:cos}) and \(Z_i\) is a normalization term reflecting the information content of the cluster.
Each merged token \(\hat{m}_i\) is stored in the merged buffer \(\mathcal{G}_u\), which maintains compact representatives of recent clusters. 
In practice, such aggregation is performed only when the evicted buffer \(\mathcal{E}_u\) accumulates enough unmatched tokens (Algorithm~\ref{alg:merge}, lines 12--19).

In the online streaming setting, each incoming token is processed on arrival: we first match it to \(\mathcal{G}_u\) and merge it when the similarity exceeds \(\lambda\); otherwise we buffer it in \(\mathcal{E}_u\) for later aggregation.
Concretely, for an incoming token \(m\), we find its nearest neighbor \(\hat m^p \in \mathcal{G}_u\) and directly absorb \(m\) when \(\cos_k(m,\hat m^p)\ge \lambda\) (Algorithm~\ref{alg:merge}, lines 3--7).
Given intra-cluster coherence, we use the matched representative \(\hat m^p\) as the reference for similarity weighting:
\begin{equation}
\hat m^p \leftarrow \frac{Z(\hat m^p)\hat m^p + \omega(\hat m^p, m)\, m}{Z(\hat m^p) + \omega(\hat m^p, m)}.
\end{equation}
Otherwise, \(m\) is appended to \(\mathcal{E}_u\) and summarized via many-to-one aggregation once the buffer is full, yielding a new representative \(\hat m_{\mathrm{new}}\).

When \(\mathcal{G}_u\) is already at capacity, inserting \(\hat m_{\mathrm{new}}\) would exceed the budget. We therefore trigger a lightweight \emph{re-merging} step that frees one slot while preserving information: it fuses the least important token with its most similar neighbor. The least important token \(\hat{m}^l\) is determined using the aggregation weight:
\begin{equation}\label{eq:low}
\hat{m}^l := \hat{m}_i^l = \mathop{\arg\min}_{\hat{m} \in \mathcal{G}_u} Z(\hat{m}),
\end{equation}
where index \(i\) indicates that \(\hat{m}_i^l\) was originally aggregated from cluster \(\mathcal{C}_i\), and its closest neighbor \(\hat{m}^p\) is chosen as:
\begin{equation}\label{eq:simi}
\hat{m}^p := \hat{m}_j^p = \mathop{\arg\max}_{\hat{m} \in \mathcal{G}_u \setminus \{\hat{m}^l\}} 
\cos_k(\hat{m}^l, \hat{m}),
\end{equation}
where index \(j\) indicates the corresponding cluster \(\mathcal{C}_j\).

% Under an online streaming setting with a fixed buffer budget, a secondary \emph{re-merging} step is triggered once \(\mathcal{G}_u\) becomes full. The step fuses the least important token with its most similar neighbor to free one slot while preserving information. The least important token \(\hat{m}^l\) is determined using the aggregation weight:
% \begin{equation}\label{eq:low}
% \hat{m}^l := \hat{m}_i^l = \mathop{\arg\min}_{\hat{m} \in \mathcal{G}_u} Z(\hat{m}),
% \end{equation}
% where index \(i\) indicates that \(\hat{m}_i^l\) was originally aggregated from cluster \(\mathcal{C}_i\), and its closest neighbor $\hat{m}^p $ is then chosen as:
% \begin{equation}\label{eq:simi}
% \hat{m}^p := \hat{m}_j^p = \mathop{\arg\max}_{\hat{m} \in \mathcal{G}_u \setminus \{\hat{m}^l\}} 
% \cos_k(\hat{m}^l, \hat{m}),
% \end{equation}
% where index \(j\) indicates the corresponding cluster \(\mathcal{C}_j\). 

In an ideal offline setting, the re-merged representation would be computed over the union of the original clusters:
\begin{align}
\hat{m}^* &= 
\frac{\sum_{m\in\mathcal{C}_i\cup\mathcal{C}_j}\omega(m^{q*},m)\,m}
     {\sum_{m\in\mathcal{C}_i\cup\mathcal{C}_j}\omega(m^{q*},m)},\\
m^{q*} &= \mathop{\arg\max}_{m\in\{m_i^q,m_j^q\}} \mathrm{score}(m).
\end{align}
However, in our streaming regime, the original tokens are {fused and discarded} once a cluster is merged, so \(\mathcal{C}_i\) and \(\mathcal{C}_j\) are no longer accessible and the ideal formulation above is infeasible. 

To derive a tractable approximation without storing all original tokens, we leverage intra-cluster coherence and adopt a \emph{small-angle approximation}:
\[
m_i \approx m_i^q \approx \hat{m}_i^l, \quad m_j \approx m_j^q \approx \hat{m}_j^p,
\]
so that the pivot, the merged token, and individual tokens in a cluster are treated as nearly aligned in feature space.

We also use \(Z(\cdot)\) as a proxy for token importance: larger \(Z(\hat{m})\) reflects greater accumulated information within the cluster. Combined with Eq.~\eqref{eq:low}, this gives the following choice of the new pivot:
\[
m^{q*} \approx \hat{m}_j^p = \mathop{\arg\max}_{\hat{m} \in \{\hat{m}_i^l, \hat{m}_j^p\}} Z(\hat{m}),
\]
from which the re-merged token is obtained as:
\begin{equation}
\begin{aligned}
\hat{m}^* &\approx \frac{
\sum_{m_j \in \mathcal{C}_j} \omega(\hat{m}_j^p, m_j)\, m_j +
\sum_{m_i \in \mathcal{C}_i} \omega(\hat{m}_j^p, m_i)\, m_i
}{
\sum_{m_j \in \mathcal{C}_j} \omega(\hat{m}_j^p, m_j) +
\sum_{m_i \in \mathcal{C}_i} \omega(\hat{m}_j^p, m_i)
} \\
&= \frac{Z_j \hat{m}_j^p + \sum_{m_i \in \mathcal{C}_i} \omega(\hat{m}_j^p, m_i)\, m_i}{Z_j + \sum_{m_i \in \mathcal{C}_i} \omega(\hat{m}_j^p, m_i)}.
\end{aligned}
\end{equation}

To further simplify, we invoke Eq.~\eqref{eq:simi} and assume \(\hat{m}_i^l \approx \hat{m}_j^p\), i.e. the two merged tokens form a small angle in feature space. Under assumptions above, the weight term can be approximated as follows.

\begin{proof}
Assume \(\hat{m}_j \approx \hat{m}_i\) and \(m_i \approx \hat{m}_i\) (omitting superscripts for clarity). We aim to show
\begin{equation} \label{eq:target}
\omega(\hat{m}_j, m_i) \approx e^{-1} \omega(\hat{m}_j, \hat{m}_i)\, \omega(\hat{m}_i, m_i).
\end{equation}
Let the normalized key vector of token \(m_i\) be $n_i = \frac{k_i}{\|k_i\|}$, so the cosine similarities in Eq.~\eqref{eq:cos} can be rewritten as:
\begin{align*}
\cos_k(\hat{m}_j, \hat{m}_i) &= \hat{n}_j \cdot \hat{n}_i, \\
\cos_k(m_i, \hat{m}_i) &= n_i \cdot \hat{n}_i.
\end{align*}
Define the small deviations
\begin{align*}
\eta = \hat{n}_j - \hat{n}_i \approx 0, \quad
\delta = n_i - \hat{n}_i \approx 0.
\end{align*}
Then
\begin{equation}
\begin{aligned}
\omega(\hat{m}_j, m_i)
&= \exp\left( \cos_k(\hat{m}_j, m_i) \right) \\
&= \exp\left( \hat{n}_j \cdot n_i \right) \\
&= \exp\left( \hat{n}_j \cdot (\hat{n}_i + \delta) \right) \\
&= \exp\left( \hat{n}_j \cdot \hat{n}_i + \delta \cdot (\hat{n}_i + \eta) \right).
\end{aligned}
\label{eq:omega-expand}
\end{equation}
Neglecting the second-order term \(\delta \cdot \eta\) and using \(\delta \cdot \hat{n}_i = n_i \cdot \hat{n}_i - 1\), Eq.~\eqref{eq:omega-expand} gives
\begin{align*}
\omega(\hat{m}_j, m_i) 
&\approx \exp\left( \hat{n}_j \cdot \hat{n}_i + n_i \cdot \hat{n}_i - 1 \right) \\
&= e^{-1}  \omega(\hat{m}_j, \hat{m}_i)\, \omega(\hat{m}_i, m_i),
\end{align*}
which proves Eq.~\eqref{eq:target}.
\end{proof}

Thus, the final re-merged token can be written as
\begin{equation}
   \begin{aligned}
\hat{m}^* &\approx \frac{Z_j \hat{m}_j^p + e^{-1} \omega(\hat{m}_j^p, \hat{m}_i^l) \sum_{m_i \in \mathcal{C}_i} \omega(\hat{m}_i^l, m_i)\, m_i}{Z_j + e^{-1} \omega(\hat{m}_j^p, \hat{m}_i^l) \sum_{m_i \in \mathcal{C}_i} \omega(\hat{m}_i^l, m_i)} \\
&= \frac{Z_j \hat{m}_j^p + e^{-1} \omega(\hat{m}_j^p, \hat{m}_i^l)\, Z_i \hat{m}_i^l}{Z_j + e^{-1} \omega(\hat{m}_j^p, \hat{m}_i^l)\, Z_i}.
\end{aligned} 
\end{equation}

\section{More Experiments}
\label{sec:supp_more_exp}

\subsection{Additional Qualitative Visualizations}
\label{sec:supp_more_vis}
\begin{figure*}[t]
    \centering
    \includegraphics[width=0.9\textwidth]{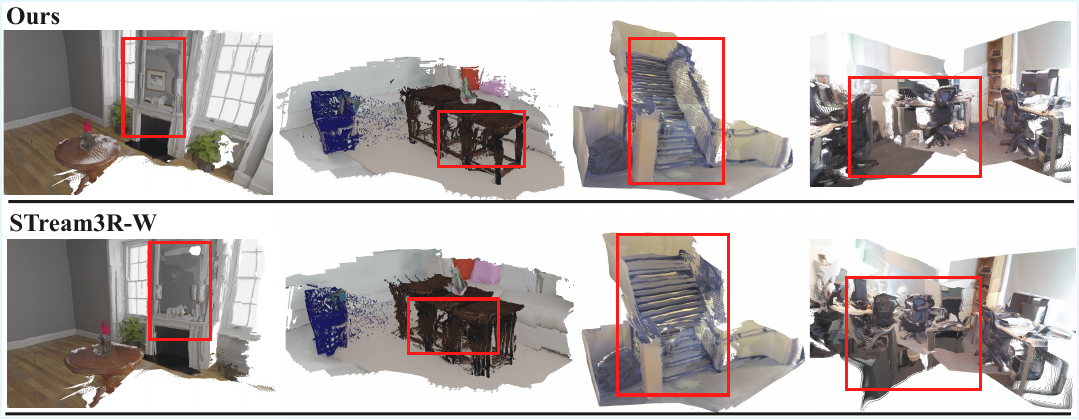}
    \caption{Additional qualitative comparisons with baselines.}
    \label{fig:supp_quality_more}
        \vspace{-8pt}

\end{figure*}

We provide additional qualitative comparisons with representative baselines.
Figure~\ref{fig:supp_quality_more} presents side-by-side reconstructions on challenging scenes.
Under the same runtime memory budget, our method preserves finer structures and yields more temporally consistent reconstructions.

\subsection{Video Depth Evaluation}
\label{sec:supp_video_depth}

\begin{table*}[t]
  \centering
  \footnotesize
\caption{
Comparison of scale-invariant depth estimation results (using per-sequence alignment) on the {Sintel}~\cite{butler2012naturalistic}, {Bonn}~\cite{palazzolo2019refusion}, and {KITTI}~\cite{geiger2013vision} datasets. 
We evaluate accuracy (Abs Rel) and consistency ($\delta < 1.25$) across different categories of methods. 
STream3R-W and StreamVGGT-W serve as sliding-window attention baselines, while our STream3R-STAC and StreamVGGT-STAC apply the cache compression strategy under the same runtime memory budget. 
Methods that rely on global alignment are marked as “GA”. 
The “Type” column indicates whether the method is optimization-based (“Optim”), streaming (“Online”), or offline (“Offline”).
}
  \label{tab:video-depth}
  \setlength{\tabcolsep}{3.5pt}
  \renewcommand{\arraystretch}{0.88}
  \resizebox{0.9\textwidth}{!}{
  \footnotesize
  \begin{tabular}{l|c|cc|cc|cc}
    \toprule
    \multirow{2}{*}{\textbf{Method}} 
    & \multirow{2}{*}{\textbf{Type}} 
    & \multicolumn{2}{c|}{\textbf{Sintel}} 
    & \multicolumn{2}{c|}{\textbf{Bonn}} 
    & \multicolumn{2}{c}{\textbf{KITTI}} \\
    &
      & \small Abs Rel~$\downarrow$ & \small ${\delta < 1.25}$~$\uparrow$
      & \small Abs Rel~$\downarrow$ & \small ${\delta < 1.25}$~$\uparrow$
      & \small Abs Rel~$\downarrow$ & \small ${\delta < 1.25}$~$\uparrow$ \\
    \midrule
    DUST3R-GA~\cite{wang2024dust3r} & Optim & 0.656 & 45.2 & 0.155 & 83.3 & 0.144 & 81.3 \\
    MASt3R-GA~\cite{leroy2024grounding} & Optim & 0.641 & 43.9 & 0.252 & 70.1 & 0.183 & 74.5 \\
    MonST3R-GA~\cite{zhang2024monst3r} & Optim & 0.378 & 55.8 & 0.067 & 96.3 & 0.168 & 74.4 \\
    \midrule
    CUT3R~\cite{wang2025continuous} & Online & 0.421 & 47.9 & 0.078 & 93.7 & 0.118 & 88.1 \\
    Spann3R~\cite{wang2024spann3r} & Online & 0.622 & 42.6 & 0.144 & 81.3 & 0.198 & 73.7 \\
    Point3R~\cite{wu2025point3r} & Online & 0.452 & 48.9 & 0.060 & 96.0 & 0.136 & 84.2 \\
    \midrule
    VGGT~\cite{wang2025vggt} & Offline & 0.297 & 68.8 & 0.057 & 96.8 & 0.061 & 96.8 \\
    \midrule
    STream3R~\cite{lan2025stream3r} & Online & \underline{0.264} & \textbf{70.5} & \underline{0.070} & \underline{95.2} & \underline{0.080} & 94.7 \\
    STream3R-W8 & Online & 0.277 & 68.9 & \textbf{0.066} & \textbf{96.5} & 0.086 & \underline{94.8} \\
    STream3R-STAC & Online & \textbf{0.259} & \underline{69.6} & 0.071 & 94.4 & \textbf{0.075} & \textbf{95.7} \\
    \midrule
    StreamVGGT~\cite{zhuo2025streaming} & Online & \textbf{0.323} & \textbf{65.7} & \textbf{0.059} & \textbf{97.2} & \textbf{0.173} & \textbf{72.2} \\
    StreamVGGT-W8 & Online & 0.349 & 62.8 & 0.068 & 96.4 & 0.201 & 67.3 \\
    StreamVGGT-STAC & Online & \underline{0.327} & \underline{65.2} & \underline{0.060} & \underline{97.1} & \underline{0.192} & \underline{67.6} \\
    \bottomrule
  \end{tabular}
  }
  \vspace{-6pt}
\end{table*}

We conduct video depth estimation by evaluating both per-frame depth quality and inter-frame consistency, aligning predicted depth maps to ground truth using a per-sequence scale following~\cite{wang2025continuous, wu2025point3r}. Experiments are conducted on three benchmark datasets—{Sintel}~\cite{butler2012naturalistic}, {Bonn}~\cite{palazzolo2019refusion}, and {KITTI}~\cite{geiger2013vision}—covering dynamic indoor and outdoor scenarios.

Results are summarized in Table~\ref{tab:video-depth}. STAC provides a \emph{training-free} mechanism for managing the KV cache under streaming input in Causal-VGGT backbones, including STream3R~\cite{lan2025stream3r} and StreamVGGT~\cite{zhuo2025streaming}. Under the same runtime memory budget, STAC consistently outperforms the sliding-window attention baselines across most datasets. Moreover, despite requiring no architectural changes or task-specific fine-tuning, it achieves accuracy comparable to full-cache Causal-VGGT models~\cite{lan2025stream3r, zhuo2025streaming} and remains robust across diverse scenes.

Furthermore, STAC preserves the representational power of the Causal-VGGT backbone, enabling it to surpass streaming-based methods such as CUT3R~\cite{wang2025continuous}, Spann3R~\cite{wang2024spann3r}, and Point3R~\cite{wu2025point3r}, as well as optimization-based approaches like DUST3R-GA~\cite{wang2024dust3r}, MASt3R-GA~\cite{leroy2024grounding}, and MonST3R-GA~\cite{zhang2024monst3r}, on most benchmarks. Notably, on Sintel, equipping STream3R with STAC outperforms the offline full-attention VGGT~\cite{wang2025vggt}, validating the effectiveness of cache compression without sacrificing model capacity.

\begin{figure*}[t]
    \centering
    \includegraphics[width=0.9\textwidth]{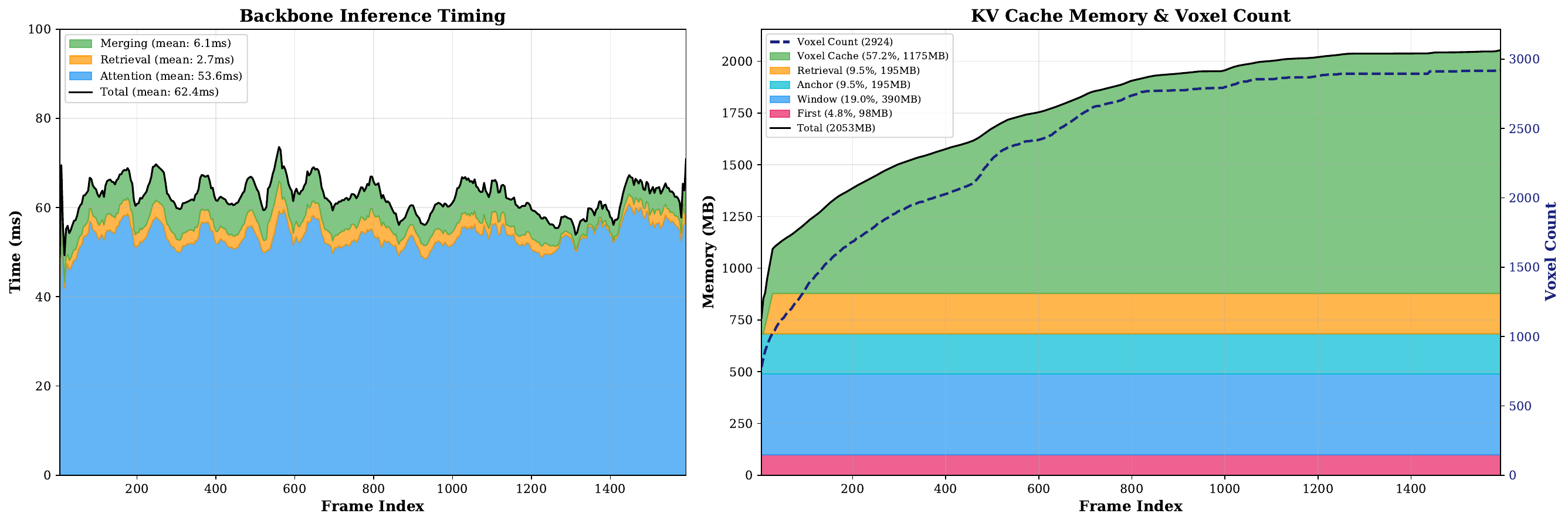}
    \caption{Behavior of STAC on a long indoor stream from the \textit{whiteroom} scene of the NRGBD dataset~\cite{azinovic2022neural}. \textbf{Left:} backbone inference time per frame with a breakdown of major components. \textbf{Right:} KV-cache memory usage and the number of active voxels (dashed, right axis) over the stream.}
    \label{fig:supp_long}
    \vspace{-10pt}
\end{figure*}

\subsection{Behavior of STAC for Long Video Streams}
\label{sec:supp_long_stream}
We analyze the behavior of STAC on a long indoor stream to verify its scalability under prolonged causal inference.
As shown in Figure~\ref{fig:supp_long}, the left panel reports the per-frame backbone inference time with a breakdown of the main components (attention+FFN, retrieval, and merging), while the right panel shows the evolution of KV-cache memory usage together with the number of voxels maintained by the long-term spatial cache.
Across the stream, the inference time remains stable (typically 70--90\,ms), indicating that retrieval and cache updates introduce a bounded overhead without accumulating latency as the sequence grows.
Meanwhile, the KV-cache memory increases only in the early stage and then gradually plateaus, and the active-voxel count converges to a stable range.
These trends suggest that STAC effectively reuses and compresses revisited spatial evidence, preventing the unbounded cache growth that would otherwise occur with full KV caching.

\subsection{Ablation of Hyperparameters}
\label{sec:supp_hyper}
\begin{table}[t]
  \centering
\caption{Effect of voxel resolution $r$ and merging threshold $\lambda$ on reconstruction quality, memory, and runtime. Experiments are conducted on the NRGBD~\cite{azinovic2022neural} dataset using Stream3R~\cite{lan2025stream3r} with stride 5. Unless otherwise specified, STAC uses the default setting $r{=}0.05$ and $\lambda{=}0.8$.}
  \label{tab:supp_hyper}
  \resizebox{1.0\columnwidth}{!}{
  \begin{tabular}{cc|ccc|cc}
    \toprule
    \textbf{$r$} & \textbf{$\lambda$}
    & \small{\textbf{ACC~$\downarrow$}}
    & \small{\textbf{Comp~$\downarrow$}}
    & \small{\textbf{NC~$\uparrow$}}
    & \small{\textbf{Mem(GB)~$\downarrow$}}
    & \small{\textbf{Time(ms)~$\downarrow$}} \\
    \midrule
    0.025 & 0.8 & 0.064 & 0.014 & 0.695 & 4.861 & 78.85 \\
    0.10  & 0.8 & 0.065 & 0.015 & 0.698 & 1.119 & 73.52 \\
    0.05  & 0.6 & 0.065 & 0.015 & 0.696 & 1.755 & 68.78 \\
    0.05  & 0.9 & 0.064 & 0.014 & 0.700 & 2.251 & 74.70 \\
    0.05  & 0.8 & 0.065 & 0.014 & 0.700 & 2.210 & 71.18 \\
    \bottomrule
  \end{tabular}
  }
\vspace{-10pt}
\end{table}

We study two key hyperparameters in STAC: the voxel grid resolution $r$ used to discretize 3D space and the merging threshold $\lambda$ that controls how aggressively evicted tokens are fused into long-term voxel representatives. Together, they determine the granularity of spatial grouping and the strength of compression, and thus trade off reconstruction quality, memory usage, and retrieval efficiency. As shown in Table~\ref{tab:supp_hyper}, $r{=}0.05$ and $\lambda{=}0.8$ provide a robust balance across these factors.

Specifically, decreasing $r$ yields a finer voxelization and activates more voxels along the stream, which increases cache capacity demand and enlarges the neighborhood set queried during retrieval, leading to higher memory and runtime costs with only marginal accuracy gains. Conversely, an overly large $r$ groups geometrically dissimilar observations into the same voxel, forcing heterogeneous tokens to share representatives and causing information loss. For the merging threshold, a smaller $\lambda$ triggers more frequent one-to-one merges, which may over-compress and degrade reconstruction quality, while a larger $\lambda$ reduces direct merging and shifts more tokens into buffered aggregation, increasing long-term cache growth pressure and retrieval overhead.

\subsection{Ablation of Compression Strategies}
\label{sec:supp_ablation}
\begin{table}[t]
  \centering
  \caption{Ablation study on 3D reconstruction performance using the NRGBD and 7-Scenes~\cite{sturm2012benchmark} datasets.
Metrics include Accuracy (ACC), Completion (Comp), and Normal Consistency (NC).
The baseline uses a sliding-window attention mechanism.
“Random Selection” selects tokens without relevance estimation; “Uniform Merging” merges tokens uniformly; and “Ours” applies attention-guided selection and weighted merging.}
  \label{tab:abl-stra}
  \resizebox{1.0\columnwidth}{!}{
  \begin{tabular}{l|ccc|ccc}
  \toprule
  \multirow{2}{*}{\textbf{Policy}} 
  & \multicolumn{3}{c|}{\textbf{NRGBD}} 
  & \multicolumn{3}{c}{\textbf{7-Scenes}} \\
  & \small{ACC~$\downarrow$} & \small{Comp~$\downarrow$} & \small{NC~$\uparrow$} 
  & \small{ACC~$\downarrow$} & \small{Comp~$\downarrow$} & \small{NC~$\uparrow$}  \\
  \midrule
  Baseline         & 0.078 & 0.015 & 0.687 & 0.107 & 0.035 & 0.587 \\
  Random Selection & 0.076 & 0.021 & 0.696 & 0.101 & 0.039 & 0.591 \\
  Uniform Merging  & 0.065 & 0.015 & 0.699 & 0.047 & 0.025 & 0.606 \\
  Ours(STAC)       & {0.065} & {0.014} & {0.700} & {0.047} & {0.024} & {0.606} \\
  \bottomrule
  \end{tabular}
  }
\vspace{-10pt}
\end{table}

To assess the effectiveness of STAC, we conduct an ablation study within STream3R~\cite{lan2025stream3r} on  NRGBD~\cite{azinovic2022neural} and 7-Scenes~\cite{sturm2012benchmark} for 3D reconstruction. Table~\ref{tab:abl-stra} compares different cache compression strategies.
\begin{itemize}\setlength{\itemsep}{0pt}
\item \textit{Selection Strategy.}
We compare token selection policies in the \emph{Working Temporal Token Cache}.
Our attention-guided policy retains anchor tokens and evicts less relevant ones for merging, whereas random selection chooses tokens indiscriminately.
The attention-guided strategy better preserves temporally persistent evidence under limited memory and outperforms random selection.

\item \textit{Merging Strategy.}
We analyze token fusion in the \emph{Long-term Spatial Token Cache}.
Uniform averaging treats all tokens equally and ignores feature relevance.
In contrast, our weighted merging better preserves informative components during fusion and yields consistently better reconstruction quality across both datasets. Notably, since tokens are grouped by voxel regions, the spatial partition itself already introduces a degree of implicit discrimination, which partially explains why uniform merging achieves relatively competitive performance.
\end{itemize}

\end{document}